\documentclass[10pt,conference]{IEEEtran}
\IEEEoverridecommandlockouts

\usepackage{hyperref}
\usepackage{cite}
\ifCLASSINFOpdf
   \usepackage[pdftex]{graphicx}
   \graphicspath{{figs/}}
   \DeclareGraphicsExtensions{.pdf,.jpeg,.png}
\else
   \usepackage[dvips]{graphicx}
   \graphicspath{{../figs/}}
   \DeclareGraphicsExtensions{.eps}
\fi
\usepackage[cmex10]{amsmath}

\usepackage{amsmath, amssymb}
\usepackage{amsthm}

\usepackage{algorithmic}
\usepackage{array}
\ifCLASSOPTIONcompsoc
  \usepackage[caption=false,font=normalsize,labelfont=sf,textfont=sf]{subfig}
\else
  \usepackage[caption=false,font=footnotesize]{subfig}
\fi
\usepackage{url}
\usepackage{booktabs}
\usepackage{caption}
\usepackage{siunitx}

\newif\iffinal
\finalfalse
\iffinal
\else
\usepackage[switch]{lineno}
\fi

\begin{document}

\title{AquaFeat: A Features-Based Image Enhancement Model for Underwater Object Detection}

\author{
    \IEEEauthorblockN{
        Emanuel C. Silva$^{1}$, Tatiana T. Schein$^{1}$, Stephanie L. Brião$^{1}$, Guilherme L. M. Costa$^{1}$, Felipe G. Oliveira$^{2}$, \\Gustavo P. Almeida$^{1}$, Eduardo L. Silva$^{1}$, Sam S. Devincenzi$^{1}$, Karina S. Machado$^{1}$  and Paulo L. J. Drews-Jr$^{1}$ \thanks{*The authors would also like to thank the CNPQ, FINEP, ANP-PRH22, FAURG and FAPERGS organizations for their research support and financial assistance.}}
    
    \IEEEauthorblockA{
        \textit{$^{1}$Centro de Ciências Computacionais (C3). Universidade Federal do Rio Grande}\\
        Rio Grande, RS, Brasil \\
        Email: \{emanuel\_silva, tatischein, stephanie.loi, 155507, gustavo.pereira.furg, eduardolawson,\\sam.devincenzi, karina.machado, paulodrews\}@furg.br
    }
    \IEEEauthorblockA{
        \textit{$^{2}$Instituto de Ciências Exatas e Tecnologia (ICET). Universidade Federal do Amazonas} \\
        Itacoatiara, AM, Brasil\\
        Email: felipeoliveira@ufam.edu.br
    }
    
}

\maketitle

\begin{abstract}
%The severe image degradation in underwater environments, such as color distortion and low visibility, impairs the performance of object detection models. While many enhancement methods improve visual quality, they are often not optimized for downstream detection tasks. In this paper, we propose AquaFeat, a novel, plug-and-play module that performs task-driven feature enhancement for underwater object detection. AquaFeat integrates a dedicated color correction stage with a multi-scale feature enhancement network that is trained end-to-end with the detector's loss function. This strategy ensures that the enhancement process is explicitly guided to refine features most relevant to the detection task. Integrated with YOLOv8, AquaFeat significantly boosts detection performance on challenging underwater datasets, achieving state-of-the-art precision and recall. By delivering substantial accuracy gains while maintaining practical processing speeds, our model provides an effective and computationally efficient solution for real-world applications like marine ecosystem monitoring and infrastructure inspection.

The severe image degradation in underwater environments impairs object detection models, as traditional image enhancement methods are often not optimized for such downstream tasks. To address this, we propose \textbf{AquaFeat}, a novel, plug-and-play module that performs task-driven feature enhancement. Our approach integrates a multi-scale feature enhancement network trained end-to-end with the detector's loss function, ensuring the enhancement process is explicitly guided to refine features most relevant to the detection task. When integrated with YOLOv8m on challenging underwater datasets, AquaFeat achieves state-of-the-art \textbf{Precision $(0.877)$} and \textbf{Recall $(0.624)$}, along with competitive mAP scores ($mAP_{0.5}$ of $0.677$ and $mAP_{0.5:0.95}$ of $0.421$). By delivering these accuracy gains while maintaining a practical processing speed of $46.5$ FPS, our model provides an effective and computationally efficient solution for real-world applications, such as marine ecosystem monitoring and infrastructure inspection.
\end{abstract}


\IEEEpeerreviewmaketitle

\section{Introduction}

Marine biology, environmental science, and resource management depend on visual understanding systems for underwater exploration and monitoring. Object detection in these cases enables applications such as species identification, debris tracking, and infrastructure inspection. However, the underwater medium presents many challenges for computer vision algorithms, primarily due to adverse conditions. More specifically, images captured in low-light underwater scenarios suffer from severe color distortion properties, light attenuation leading to low luminosity, and scattering effects that cause blurring and reduced contrast~\cite{ye2025yes,chen2024underwater}. These degradations limit object visibility and overall image quality, consequently degrading the performance of object detection models. %Works have emerged that propose feature-based enhancement methods plugged into detection methods. 

Visual enhancement models, although seeking to mitigate these problems, often incur a high computational cost and longer execution times through deep learning (DL)~\cite{rasheed2025advancing,zhao2025deep}. This occurs because many of these methods are optimized to improve overall visual appearance, rather than optimizing specific features for downstream tasks, such as object detection. The effectiveness of object detection is not always correlated with perceived visual enhancement; an output that appears better to the human eye may not necessarily yield better detection results if key features are obscured or distorted by the enhancement process.

In this paper, we propose the AquaFeat model, which focuses on feature-level enhancement for underwater image enhancement (UIE) in low-light conditions to improve object detection performance. We incorporate color correction modifications to handle the unique chromatic distortions of the underwater environment~\cite{colorcorrectionANDSpecialConv, lai2025color}. As a foundation for our work, we use the YOLOv8m model \cite{yoloV8}, initially pre-trained on the DeepFish \cite{deepfish} and OzFish \cite{ozfish2020} datasets for the underwater domain, and subsequently fine-tuned on the FishTrach23 dataset \cite{dawkins2024fishtrack23}. Furthermore, the performance of this model was compared with visual and feature enhancement methods, which are integrated with YOLOv8m. For quantitative evaluation, we used the following metrics: mean Average Precision (mAP), with two settings: $mAP_{0.5}$ and  $mAP_{0.5:0.95}$, as well as Precision, Recall, and FPS. Thus, through analysis, we demonstrate that our approach is computationally efficient compared to underwater enhancement methods. Additionally, AquaFeat improves object detection accuracy compared to feature-based methods. The main contributions of this work are summarized as:
\begin{itemize}
    \item We propose AquaFeat, a novel plug-and-play module designed to enhance hierarchical features and boost vision tasks in low-light underwater conditions. AquaFeat is flexible and can be trained end-to-end for various high-level underwater vision applications;
    
    \item To the best of our knowledge, this is the first plug-and-play approach that enhances underwater images at the feature level before performing tasks such as object detection, semantic segmentation, and video object detection;
    
    \item We introduce a processed version of the FishTrack23 dataset, created by systematically sampling the original video data. This adaptation was necessary to curate a focused and manageable image-based dataset suitable for object detection tasks.
\end{itemize}

\section{DL-based Image Enhancement Works}

Deep learning has driven significant advances in underwater image enhancement, inspiring numerous research efforts in this field \cite{chen2024underwater}. Existing methods can be broadly categorized into two main approaches: (A) Visual-based approaches that focus on enhancing the visual quality of underwater images, and (B) Feature-based approaches that focus on enhancing features relevant for downstream tasks, such as object detection and classification.

\subsection{Visual-based Underwater Image Enhancement}

Supervised methods train networks to map degraded underwater images to clear references, often using synthesized data. For instance, a deep Retinex network~\cite{ji2023deep} was proposed for low-light conditions, featuring sub-networks for decomposition, reflectance enhancement, and illumination adjustment. Another example is UICE-MIRNet \cite{guo2025underwater}, which adapts the MIRNet architecture by adding a block to enhance colorfulness and prevent over-brightness, thereby demonstrating its utility for object detection. In contrast, unsupervised methods eliminate the need for paired data. These include Generative Adversarial Networks (GANs) like FUnIE-GAN \cite{islam2020fast} and zero-reference approaches such as Zero-UAE \cite{liu2024lightweight}, which uses a lightweight network trained with non-reference losses to estimate parameters for an adaptive curve model. Diffusion-based models are also prominent. UDBE \cite{schein2025udbe} uses a U-Net architecture guided by brightness to clarify images without reference data. The Osmosis method \cite{nathan2024osmosis} adopts a novel approach by learning a joint diffusion prior over both color and depth, thereby inverting the physical image formation process. UDnet \cite{saleh2022adaptive} is an adaptive method that automatically generates reference maps by adjusting for contrast, saturation, and gamma based on data uncertainty, allowing it to learn from limited data without annotations.

\subsection{Features-based Image Enhancement}

Recent research has explored various fronts to overcome the challenges of image degradation in underwater object detection. A promising approach is end-to-end, task-centric learning, such as that of the FeatEnHancer model \cite{hashmi2023featenhancer}. Although developed for terrestrial environments, its principle of optimizing features for the final task's metric, rather than for human visual appeal, is directly applicable to the underwater domain. Other works focus on specialized architectures. The Multi-Scale Feature Enhancement (MSFE) method \cite{li2025multi}, for example, was designed to preserve fine-grained details in blurry underwater images, while the PDSC-YOLOv8n model \cite{ding2024lightweight} adapts the popular YOLOv8 network with pre-processing algorithms and optimized convolutions for low-light scenarios. In a more integrated line of work, the AMSP-UOD network \cite{zhou2024amsp} combines image formation models with neural network optimization to mitigate multiple degradation factors, demonstrating high performance in detecting dense targets. However, despite these advances, image enhancement and feature representation-based detection are still often treated as independent stages. To overcome this gap, we introduce the methodology outlined below.

\section{AquaFeat - Proposed Model}

%Our method changes the FeatEnHancer original structure by adding a Color Correction module from MSDC-Net \cite{colorcorrectionANDSpecialConv}, a mathematical block that aims to reduce color deviation in the underwater environment in a way it does not require training [more explanation], right at the beginning of the enhancement. The next change to the original architecture is the substitution of the first layer of the Feature Enhancement Network (FEN) and the networks' last convolutional layer by the SpecialConv layer by \cite{colorcorrectionANDSpecialConv}.%The Color Correction Module works by [explanation]. While the SpecialConv layer aims to [explanation].

\begin{figure*}%[htb]
\centering
\includegraphics[width=\textwidth]{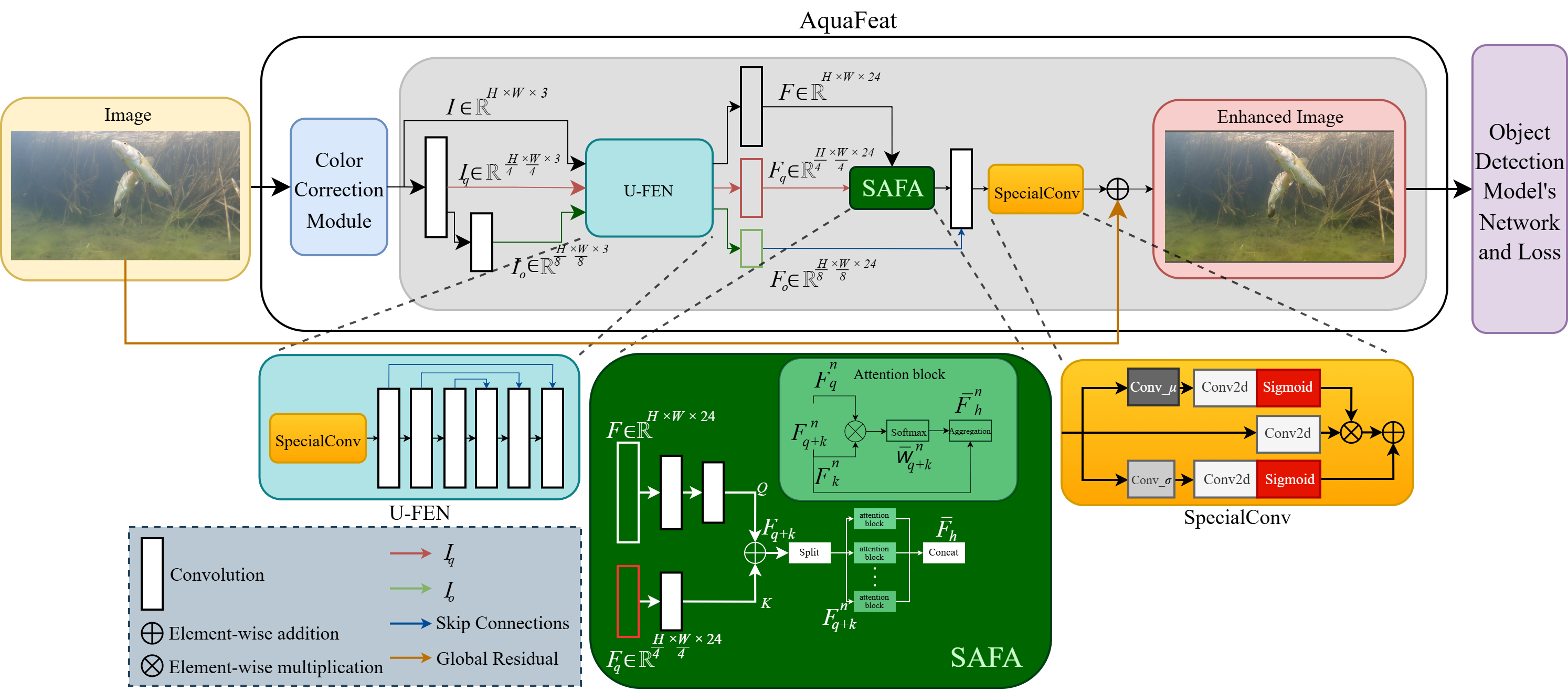}
%...\DeclareGraphicsExtensions.
\caption{Overview of the AquaFeat architecture. The pipeline begins with the color correction module, followed by the Underwater-Feature Enhancement Network (U-FEN) for feature extraction. The Scale-Aware Feature Aggregation (SAFA) module then merges these features to produce an enhanced image, which is subsequently fed into a detection network (YOLOv8m).}
%\caption{The AquaFeat architecture. After initial color correction, the Underwater-Feature Enhancement Network (U-FEN) extracts features, which are then merged by the Scale-Aware Feature Aggregation (SAFA) module to produce an enhanced image for a detection network like YOLOv8m.}
%\caption{An overview of the proposed AquaFeat model architecture. An input image first undergoes pre-processing by a Color Correction module. Then, a multi-scale Underwater-Feature Enhancement Network (U-FEN) extracts hierarchical features. These features are merged by the Scale-Aware Feature Aggregation (SAFA) module. The final enhanced image is generated via an adaptive residual connection and passed to a downstream detection network, such as YOLOv8m.}

\label{fig:aquafeat_arch}
\end{figure*}

The AquaFeat model is a plug-and-play feature enhancement approach designed to improve object detection performance in low-light underwater conditions (see Fig. \ref{fig:aquafeat_arch}). This proposed model aims to correct domain-specific degradations and explicitly enhance task-specific features, enabling downstream models to achieve higher accuracy and robustness. Our approach is implemented and evaluated using the YOLOv8m model for the downstream detection. AquaFeat consists of three main components: i) a Color Correction Module; ii) a Feature Enhancement Process; iii) an adaptive Residual Output, which are described below.  % Figure~\ref{fig:aquafeat_arch} presents an overview of the proposed AquaFeat framework.

\subsection{Color Correction Module}

This model addresses the unique chromatic distortions of the underwater environment by balancing color channel histogram distributions and compensating for wavelength-dependent light absorption. As a fundamental step, the input image undergoes a non-trainable pre-processing to correct the dominant color cast. The color correction module, based on Liu et al.~\cite{colorcorrectionANDSpecialConv}, analyzes the mean intensities of the R, G, and B channels and adjusts the color distribution towards the intensity of the median channel. This initial white balancing reduces chromatic distortions caused by varying water conditions, allowing subsequent layers to focus on more complex feature extraction. Thus, the primary objective is to mitigate the severe visual degradation inherent in underwater images, thereby providing a cleaner and more feature-rich input to subsequent vision models for tasks such as object detection, segmentation, and video analysis. 

\subsection{Feature Enhancement Process} 

This process, adapted from the FeatEnHancer methodology \cite{hashmi2023featenhancer}, integrates seamlessly with downstream vision tasks to enhance hierarchical features at multiple levels. It is divided into three stages, as described below.

\subsubsection{Underwater-Feature Enhancement Network (U-FEN)}

After performing color correction, we process the RGB image $I \in \mathbb{R}^{H \times W \times C}$ in three parallel streams at its original, quarter ($I_{q} \in \mathbb{R}^{\frac{H}{4} \times \frac{W}{4} \times C}$), and one-eighth ($I_{o} \in \mathbb{R}^{\frac{H}{8} \times \frac{W}{8} \times C}$) resolutions. Each of these streams is fed into a shared-weight Underwater-Feature Enhancement Network (U-FEN), which is inspired by image enhancement methods for adverse conditions. A key innovation in our U-FEN is the replacement of the first conventional layer with a \texttt{SpecialConv} layer \cite{colorcorrectionANDSpecialConv}. Unlike a standard convolution, \texttt{SpecialConv} introduces a trainable, content-aware mechanism that dynamically adjusts contrast. It computes channel-wise statistics (mean and standard deviation) from its input to generate adaptive multipliers, which are then applied to the convolutional features to enhance their distinctiveness. Following this layer, a \texttt{LeakyReLU} activation is applied, which helps prevent vanishing gradients while processing the features. The rest of the U-FEN encoder uses six standard convolutional layers ($3 \times 3$ kernel) with dense skip connections to effectively process and enhance features at each scale.

\subsubsection{Scale-Aware Fusion}

The Scale-Aware Feature Aggregation (SAFA) module provides a sophisticated method for fusing features from different scales, which is crucial for handling complex visual scenes. This module operates on both full-resolution ($F$) and quarter-resolution ($F_q$) feature maps. To begin, it addresses the challenge of differing spatial dimensions by projecting both inputs to a common, smaller resolution. This is achieved using distinct convolutional pathways to generate \textit{query} ($Q$) and \textit{key} ($K$) tensors. Using separate, non-shared weights for these projections allows the network to learn optimized and scale-specific embeddings before the core attention mechanism is applied. Once the features are projected, the $Q$ and $K$ tensors are concatenated, and their channel dimension is partitioned into eight parallel attention heads. This multi-head design enables the module to jointly attend to information from different representation subspaces simultaneously. Within each head, attention weights are computed and then normalized using a softmax function. These weights guide the process of creating a weighted summation of the features from the different scales, resulting in a single, contextually rich feature map that combines the most salient information. %To conclude the process, this aggregated map is upsampled to its original resolution via bilinear interpolation and is added back to the initial high-resolution feature stream through a residual connection. This final step skillfully integrates the learned multi-scale context while ensuring that fine-grained details from the original representation are preserved.

\subsubsection{Final Feature Aggregation}

To complete the feature fusion, the output from the SAFA module is integrated with the features from the smallest (eighth) resolution path. The small-scale feature map is first upsampled via bilinear interpolation to match the spatial dimensions of the SAFA output. These two feature streams are then concatenated along the channel dimension, creating a unified tensor that combines broad contextual information with fine-grained details. This tensor is immediately processed by a final $3x3$ convolutional layer, which learns to cohesively merge the features into a single, comprehensive representation before being passed to the adaptive output stage.

%Following feature extraction in the encoder, the information from the different scales must be intelligently merged. This is the role of the \textbf{Scale-Aware Feature Aggregation (SAFA)} module. Instead of a simple concatenation, SAFA uses a learned attention mechanism to fuse the main feature stream with the quarter-scale feature stream. This allows the network to integrate broad, contextual information with fine-grained local details in a content-dependent manner, producing a highly informative, unified feature map that is passed to the decoder path.

\subsection{Adaptive Residual Output}

The final stage of the AquaFeat model is responsible for generating the enhanced output image from the aggregated feature map. This stage begins with a final \texttt{SpecialConv} layer. Crucially, this \texttt{SpecialConv} instance serves a different purpose than the one in the U-FEN encoder. Its role here is not intermediate feature extraction, but the generation of a final enhancement residual — a map of the corrections needed for the image. To ensure model stability, the output of this layer is passed through a \texttt{tanh} activation function. The \texttt{tanh} function constrains the enhancement values to a fixed range of $[-1, 1]$, preventing extreme adjustments that could destabilize the training process. This bounded residual is then added back to the original, unmodified input image. The loss function depends on the downstream task, which in this case is object detection with YOLOv8m~\cite{yoloV8}. The model parameters are optimized through task-related loss functions, enabling the enhancement network to improve only task-relevant features rather than relying on generic image enhancement losses. This final residual connection is a stable and effective strategy for producing the feature-rich image that is subsequently passed to the chosen task.

\section{Experimental Results}

This section presents both a qualitative and quantitative evaluation of the proposed approach, comparing it with visual enhancement techniques and feature enhancement methods associated with YOLOv8m for object detection.

\subsection{Datasets}

For this work, the YOLOv8m model was pre-trained using two publicly available Australian datasets: OzFish \cite{ozfish2020}, containing approximately 43,000 annotations across 1,800 images with high fish density (average of 25 fish per frame), and DeepFish \cite{deepfish}, with around 15,000 annotations in 4,505 images from diverse habitats. The model was then fine-tuned on the FishTrack23 dataset \cite{dawkins2024fishtrack23}, a challenging video collection with $\sim 850,000$ annotations featuring large schools of fish, dynamic backgrounds, and significant scale variations. A notable difficulty in this dataset is the low visibility of targets in certain frames, where objects are virtually imperceptible despite being annotated. We sampled one annotated frame every 30 frames and unified all labels into a single "Fish" class, resulting in $6,392$ images split into $4,474$ for training, $1,278$ for testing, and $640$ for validation.

\subsection{Implementation Details}

To ensure a consistent and fair comparison, all experiments were conducted on a single workstation. The hardware configuration consisted of an AMD Ryzen 7 8700G CPU, 32 GB of DDR5-4800 RAM, and an NVIDIA GeForce RTX 4070 Ti GPU with 12 GB of GDDR6X VRAM. Our proposed model, AquaFeat, which utilizes YOLOv8m as its detection backbone, was trained using the AdamW optimizer. The final hyperparameters were set to a batch size of 6 and a learning rate of $3 \times 10^{-4}$. 

\subsection{Comparison and Metrics} 

Our feature-based enhancement method, AquaFeat, is compared against well-established techniques in the literature, including YOLOv8m \cite{yoloV8} and YOLOv10s \cite{YOLOv10} as a baseline, FeatEnHancer \cite{hashmi2023featenhancer}, which performs feature-level enhancement for low-light terrestrial scenarios, and AMSP-UOD \cite{zhou2024amsp}, which also utilizes feature-based enhancement for underwater object detection. In addition, we evaluate AquaFeat against traditional visual enhancement methods, such as OSMOSIS \cite{nathan2024osmosis}, UDNET \cite{saleh2022adaptive}, and UDBE \cite{schein2025udbe}, all of which are integrated with YOLOv8m. To ensure a fair comparison, we strictly follow the training protocols and hyperparameter settings reported in the original publications for each method. To evaluate and compare the results we use the metrics $mAP_{0.5}$, $mAP_{0.5:0.95}$, Precision \cite{manning2009introduction}, Recall \cite{manning2009introduction}.In addition to the accuracy metrics, the system's temporal performance, measured in frames per second (FPS), was also evaluated. We aim to comprehensively assess the effectiveness of our method in comparison to these literature techniques.

%A key feature of AquaFeat is its ability to optimize enhanced image representations based on the loss function of the downstream task, resulting in more expressive and task-specific features. AquaFeat fuses multi-scale hierarchical information to produce representations well-suited for underwater environments, and its parameters can be fine-tuned directly for specific applications such as underwater object detection.

\subsection{Qualitative Evaluation}

\begin{figure*}%[htb]
    % Reduces the space between the last image row and the caption text.
    \captionsetup{skip=4pt}
    \centering
    
    % Set column separation to zero to remove horizontal space between images
    \setlength{\tabcolsep}{0pt}
    
    % The column definition is updated to remove space between image columns.
    % The '@{}' tells LaTeX to insert nothing between these columns.
    \begin{tabular}{
        >{\centering\arraybackslash}m{2.2cm} % For the text labels
        >{\centering\arraybackslash}m{0.19\textwidth} % For the image columns
        @{} % This removes the space between the 2nd and 3rd image columns
        >{\centering\arraybackslash}m{0.19\textwidth}
        @{} % This removes the space between the 3rd and 4th image columns
        >{\centering\arraybackslash}m{0.19\textwidth}
        @{} % This removes the space between the 4th and 5th image columns
        >{\centering\arraybackslash}m{0.19\textwidth}
    }
    % Row 1: Ground Truth
    \scriptsize Ground Truth &
    \includegraphics[width=\linewidth]{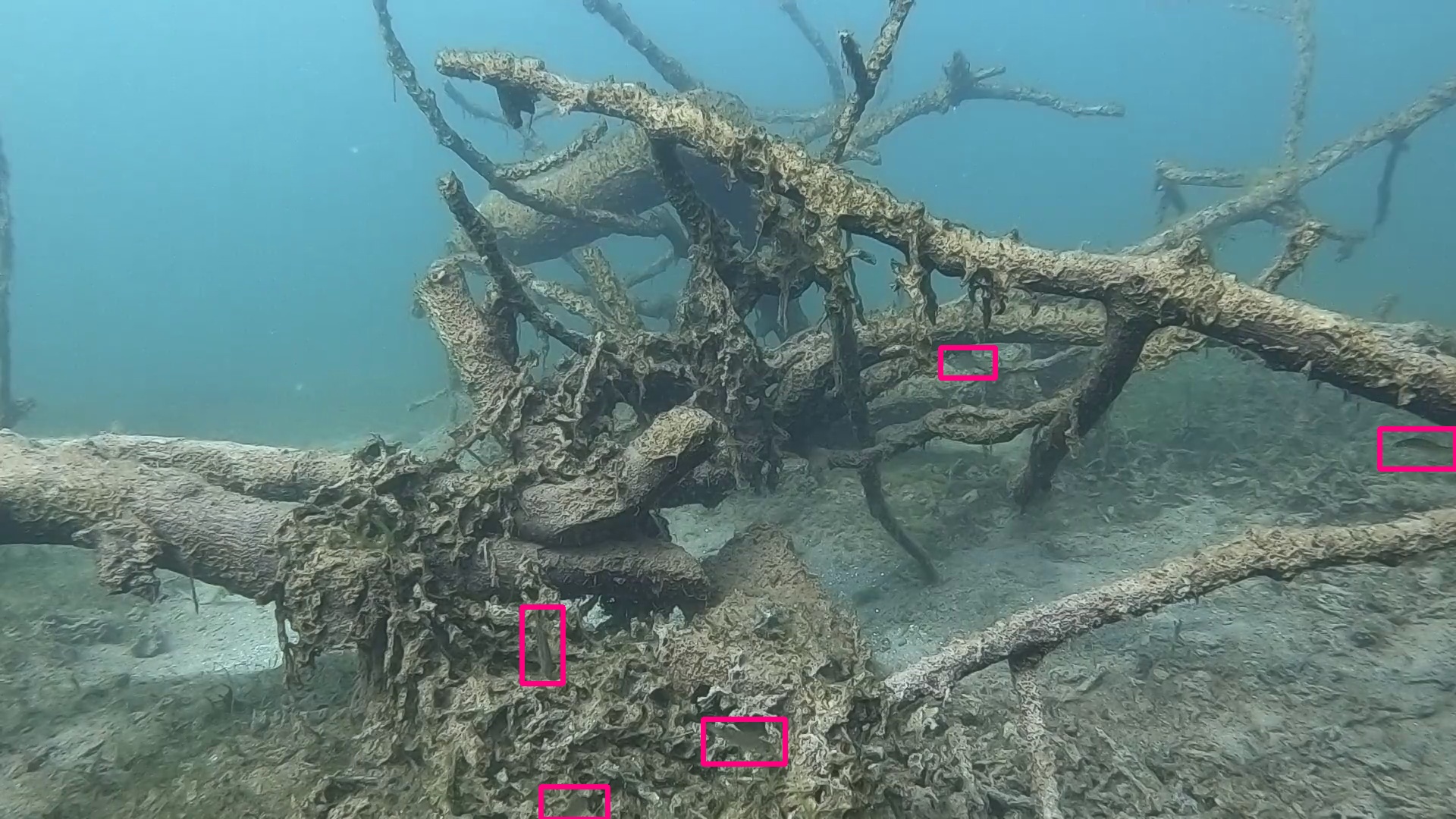} &
    \includegraphics[width=\linewidth]{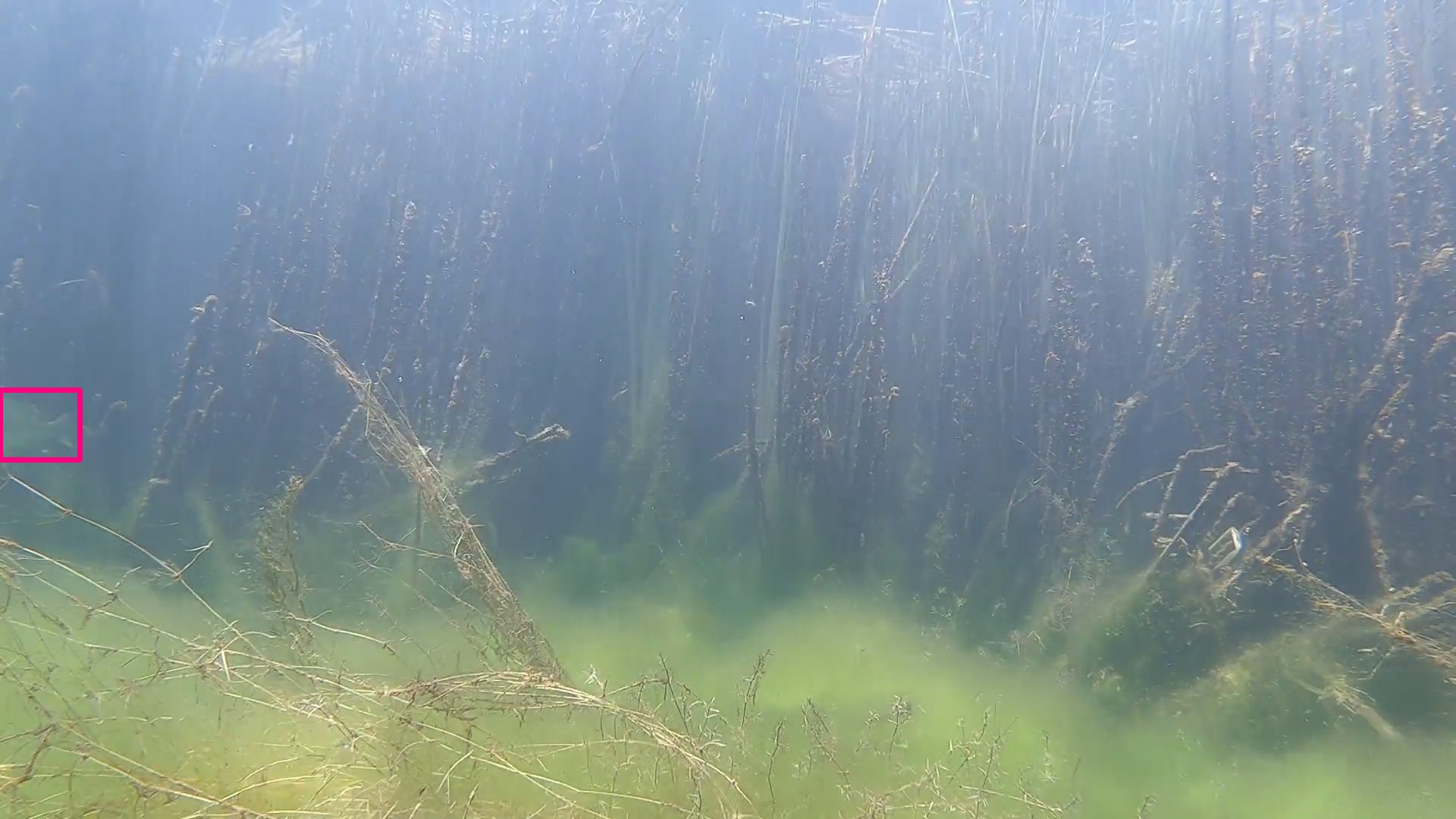} &
    \includegraphics[width=\linewidth]{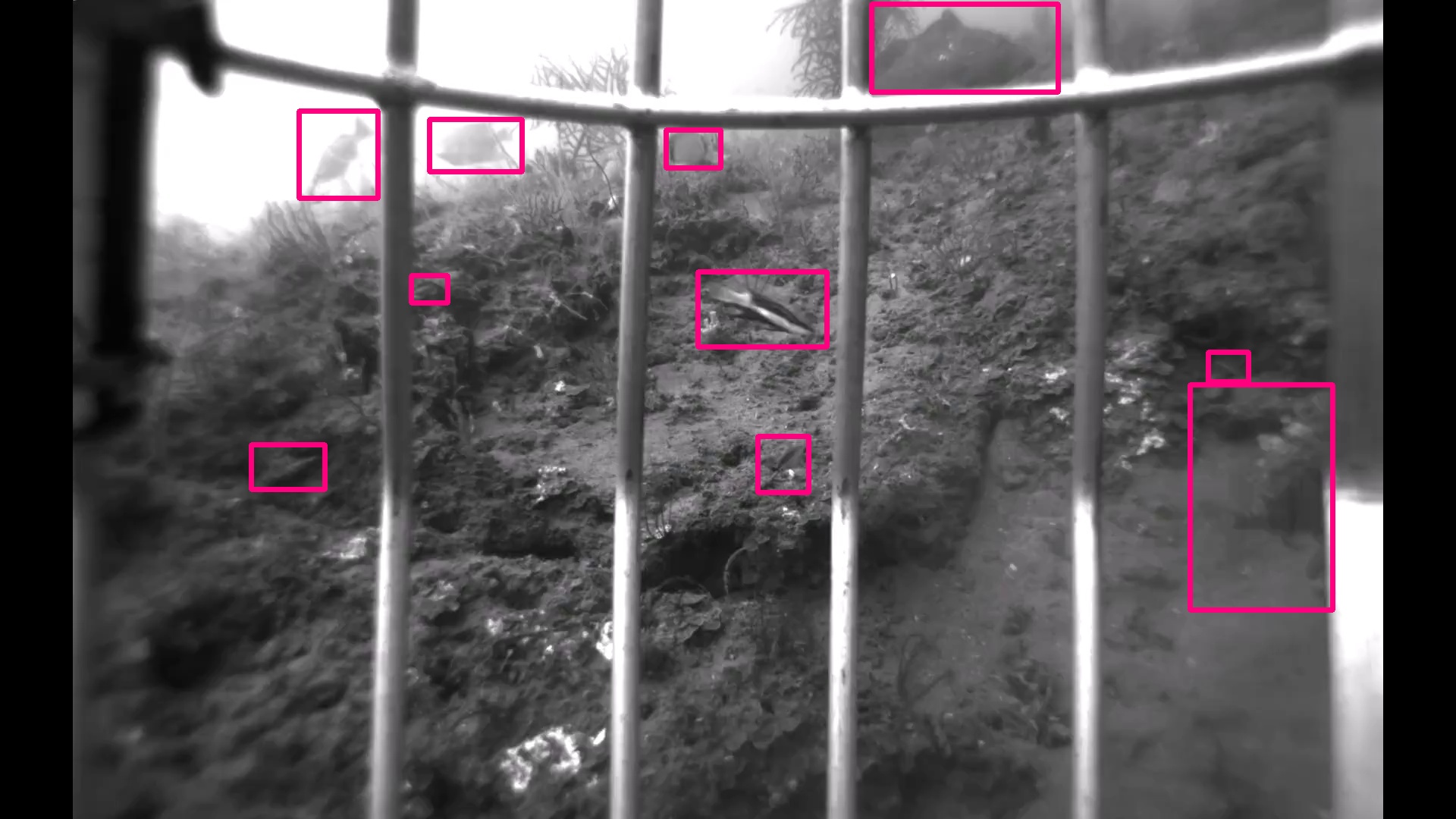} &
    \includegraphics[width=\linewidth]{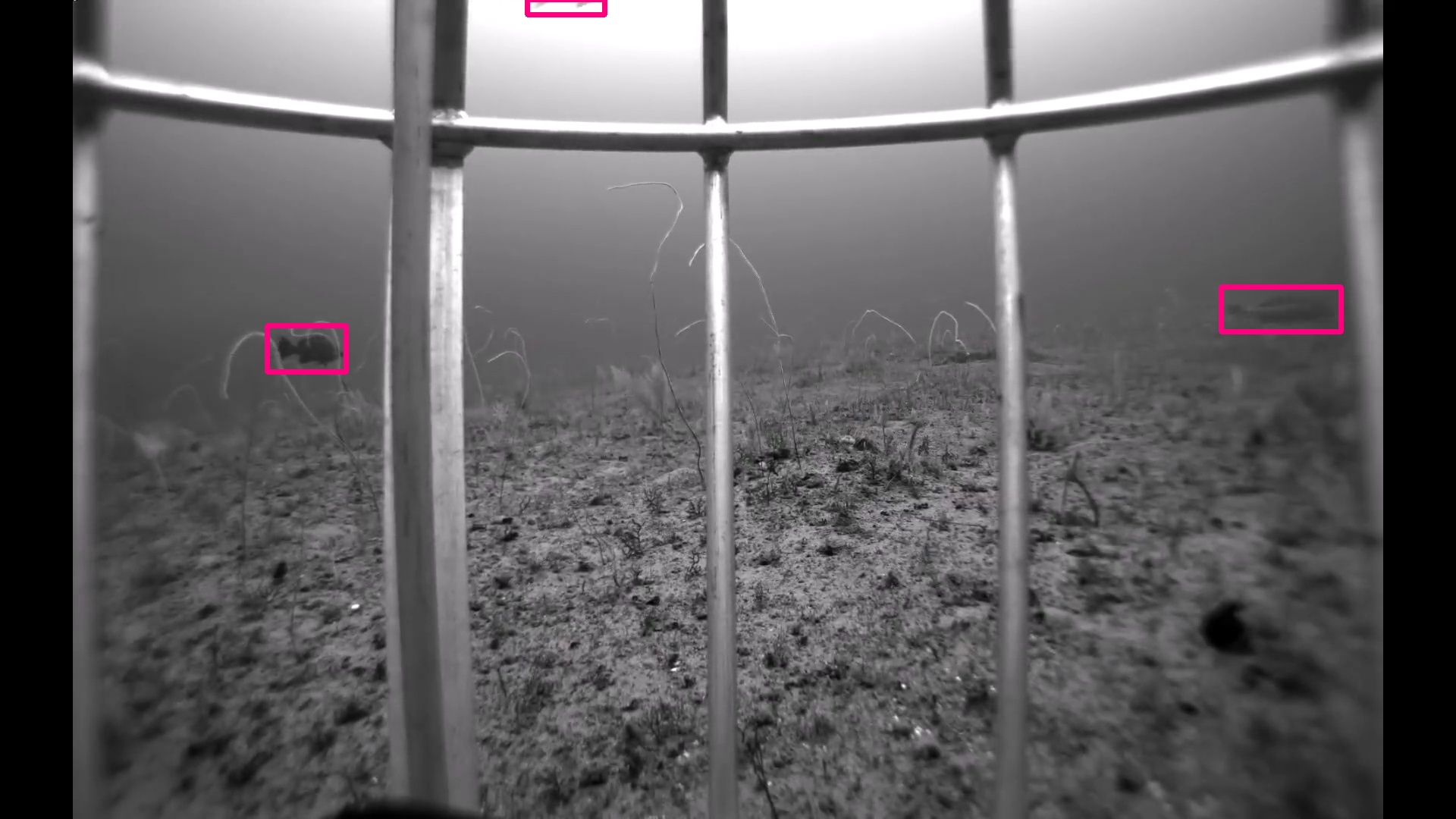} \\[-4pt] % Reduces space after this row

    % Row 2: FeatEnHancer
    \scriptsize FeatEnHancer \cite{hashmi2023featenhancer} &
    \includegraphics[width=\linewidth]{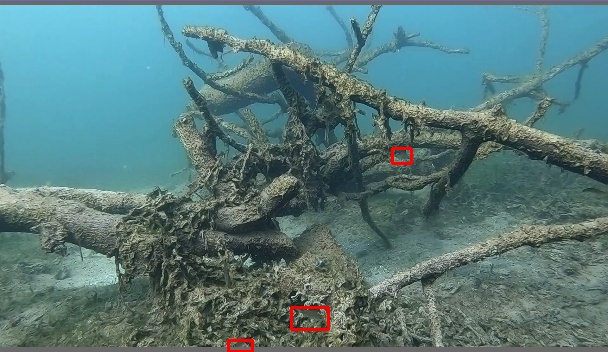} &
    \includegraphics[width=\linewidth]{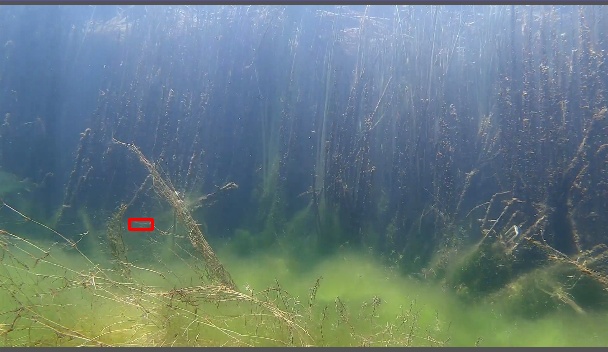} &
    \includegraphics[width=\linewidth]{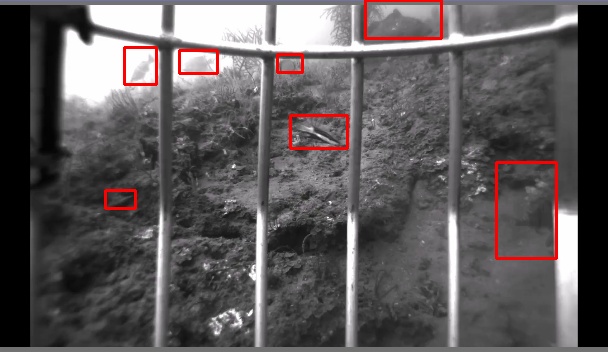} &
    \includegraphics[width=\linewidth]{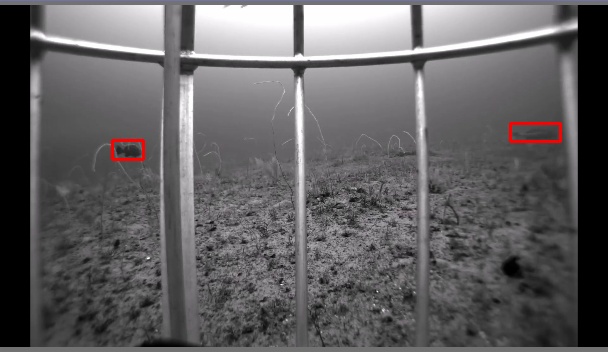} \\[-4pt]

    % Row 3: AMSP
    \scriptsize AMSP - UOD \cite{zhou2024amsp} &
    \includegraphics[width=\linewidth]{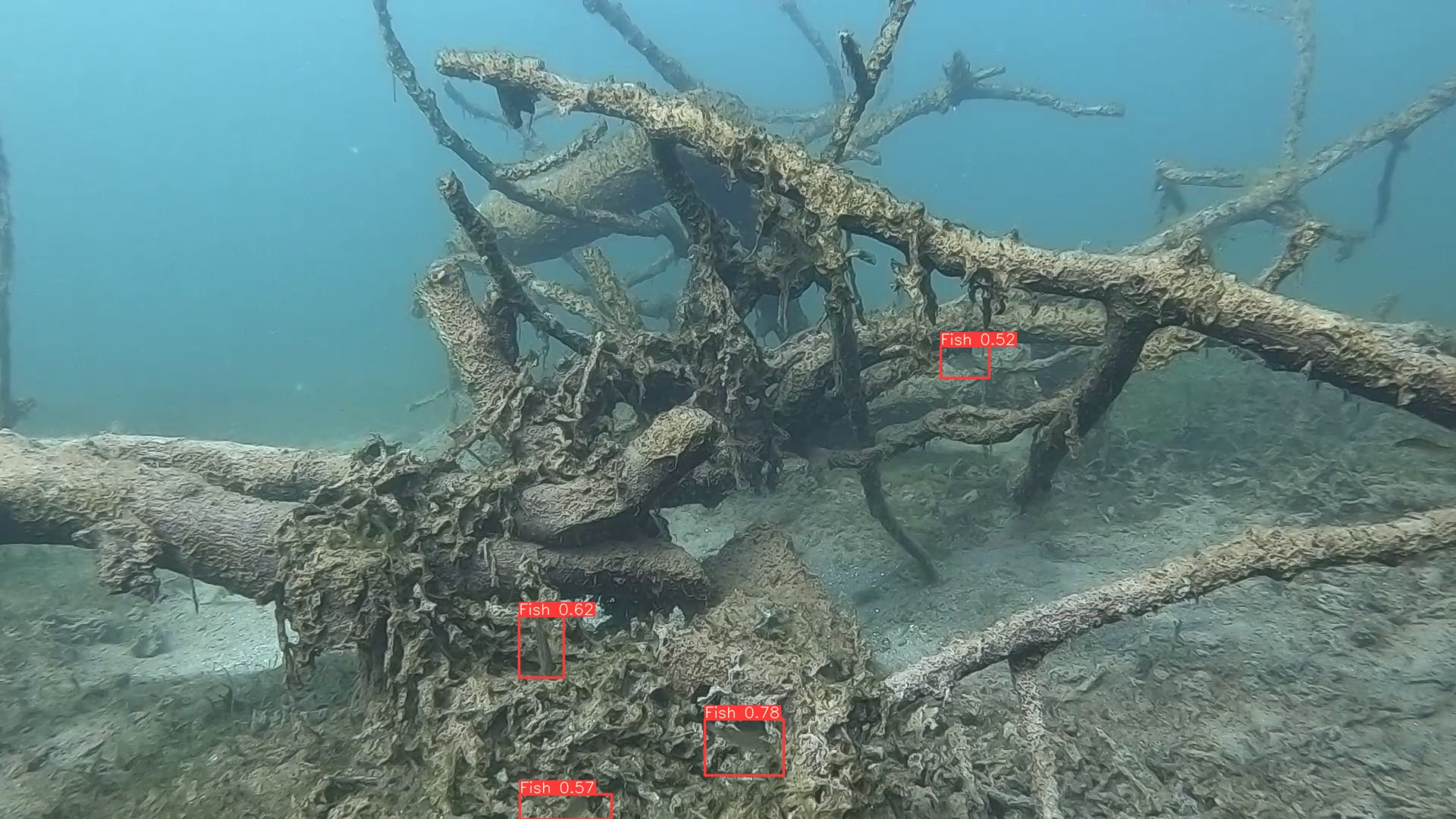} &
    \includegraphics[width=\linewidth]{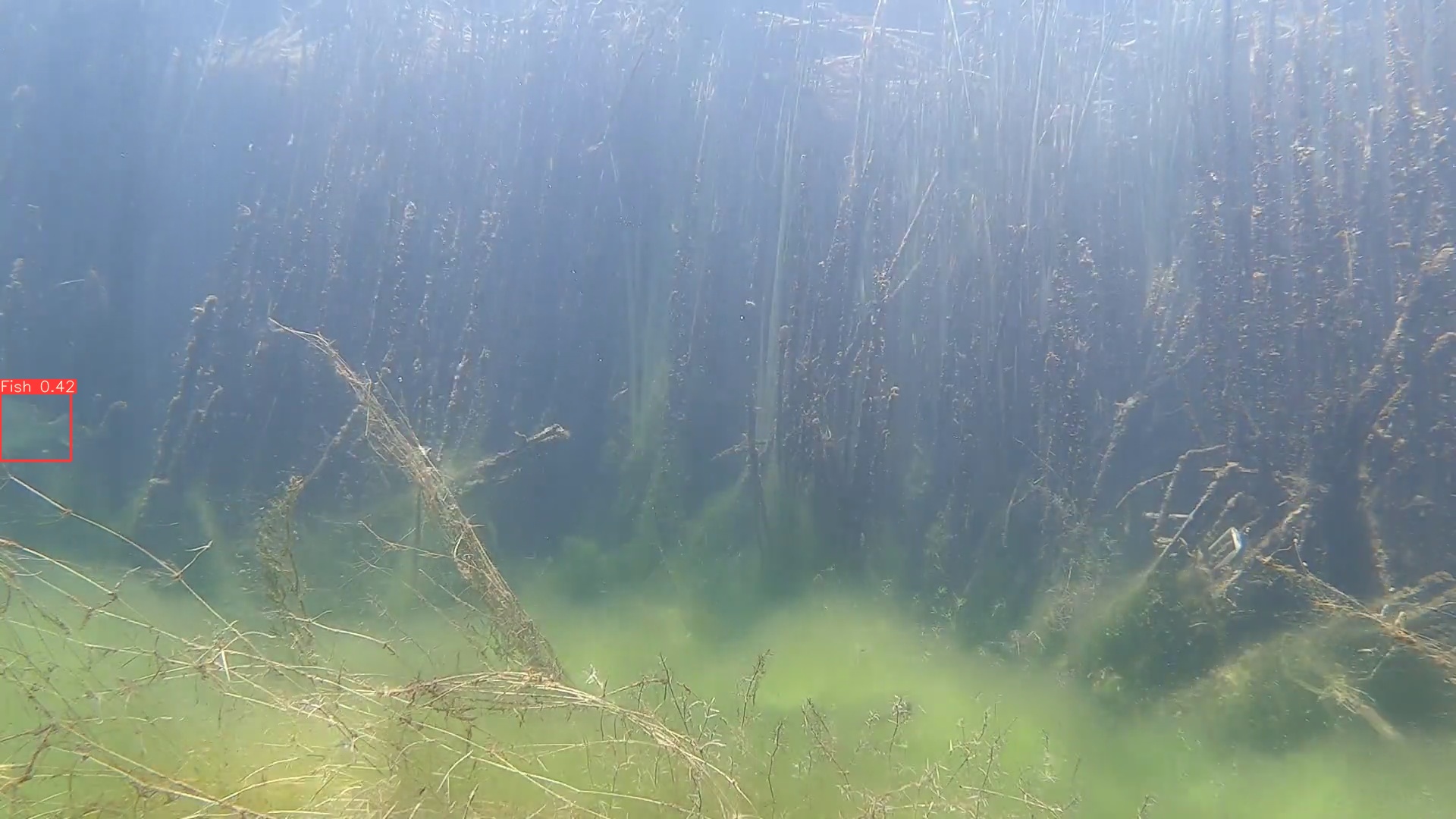} &
    \includegraphics[width=\linewidth]{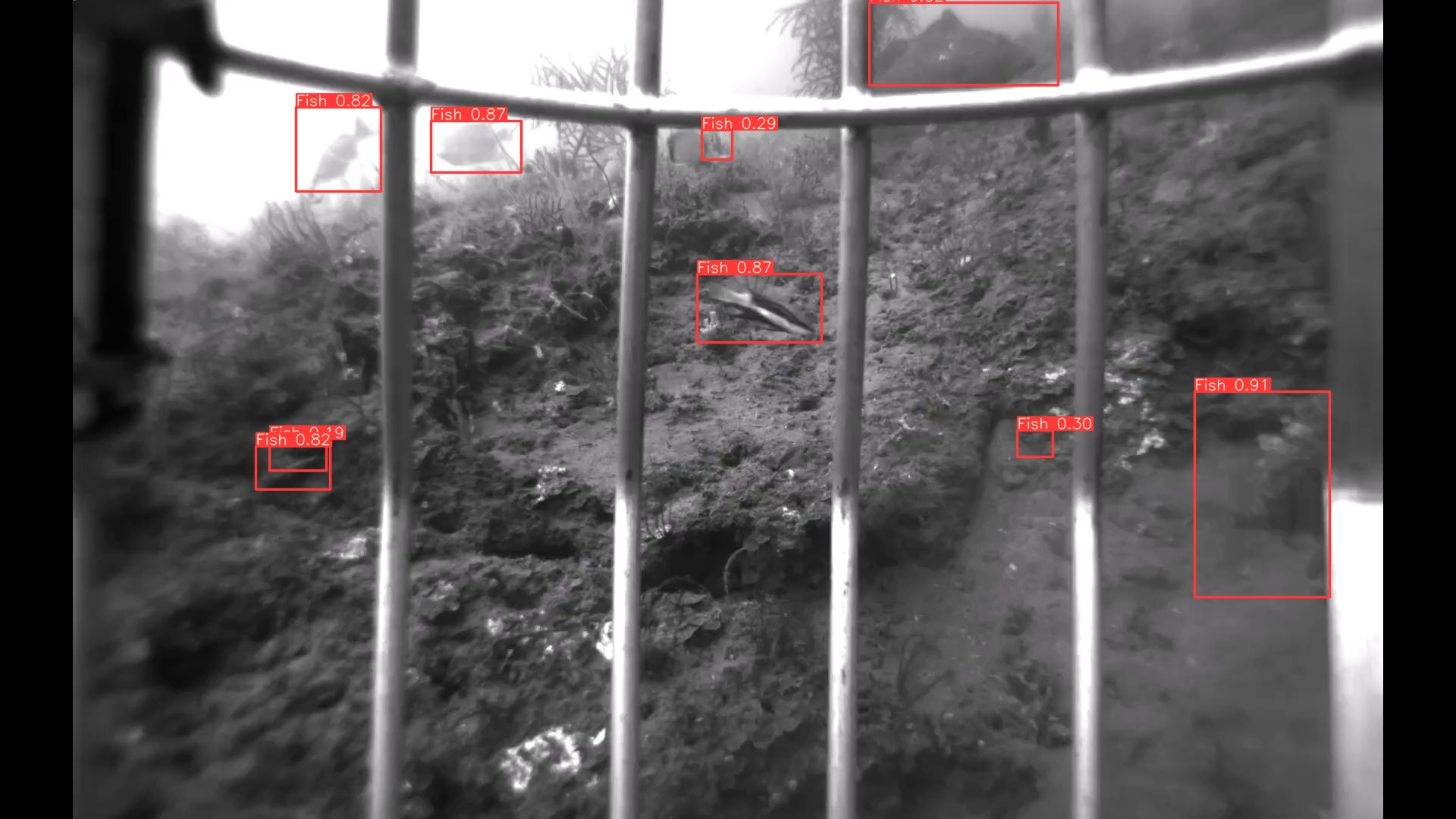} &
    \includegraphics[width=\linewidth]{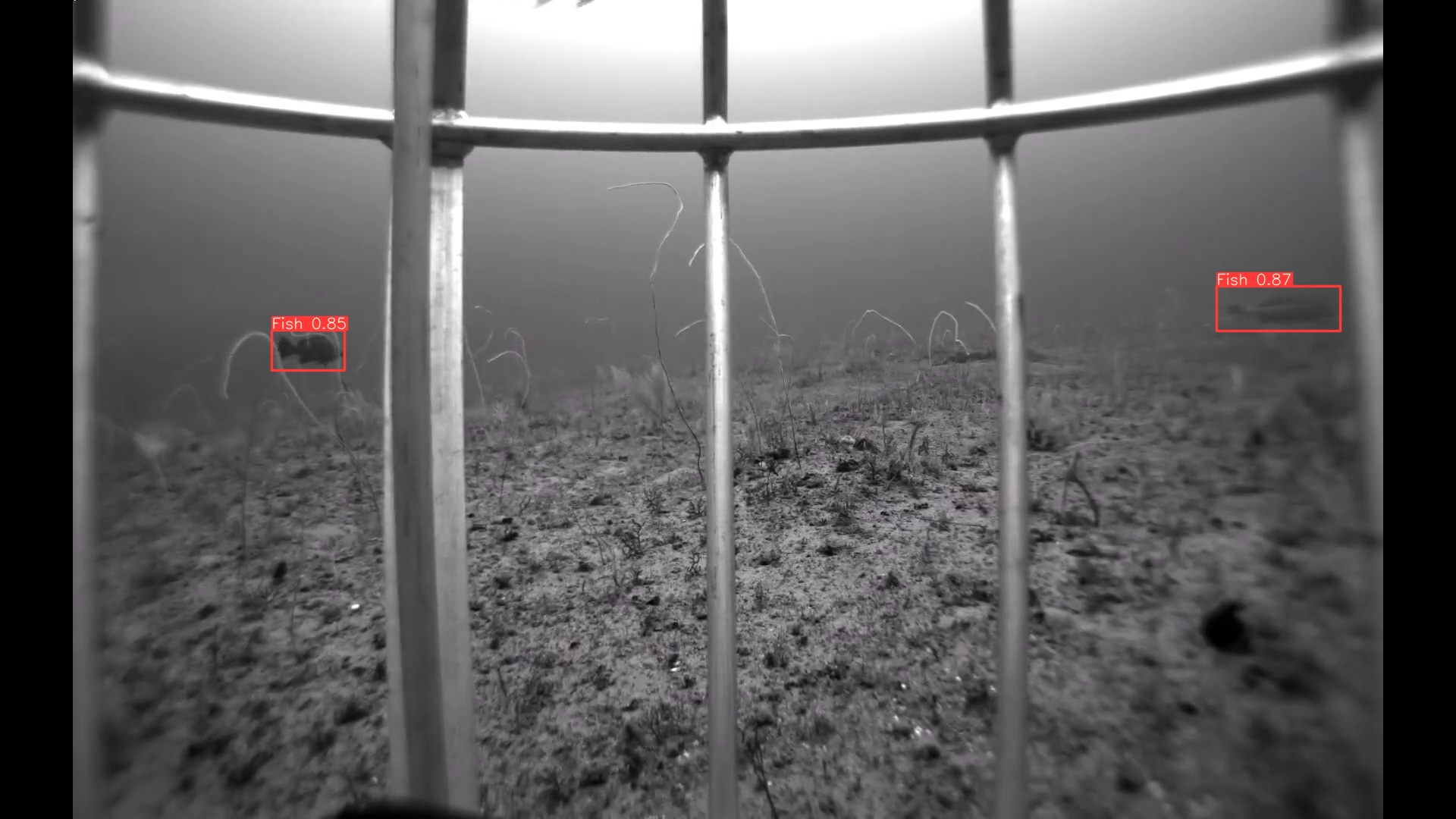} \\[-4pt]

    % Row 4: Osmosis
    \scriptsize Osmosis \cite{nathan2024osmosis} &
    \includegraphics[width=\linewidth]{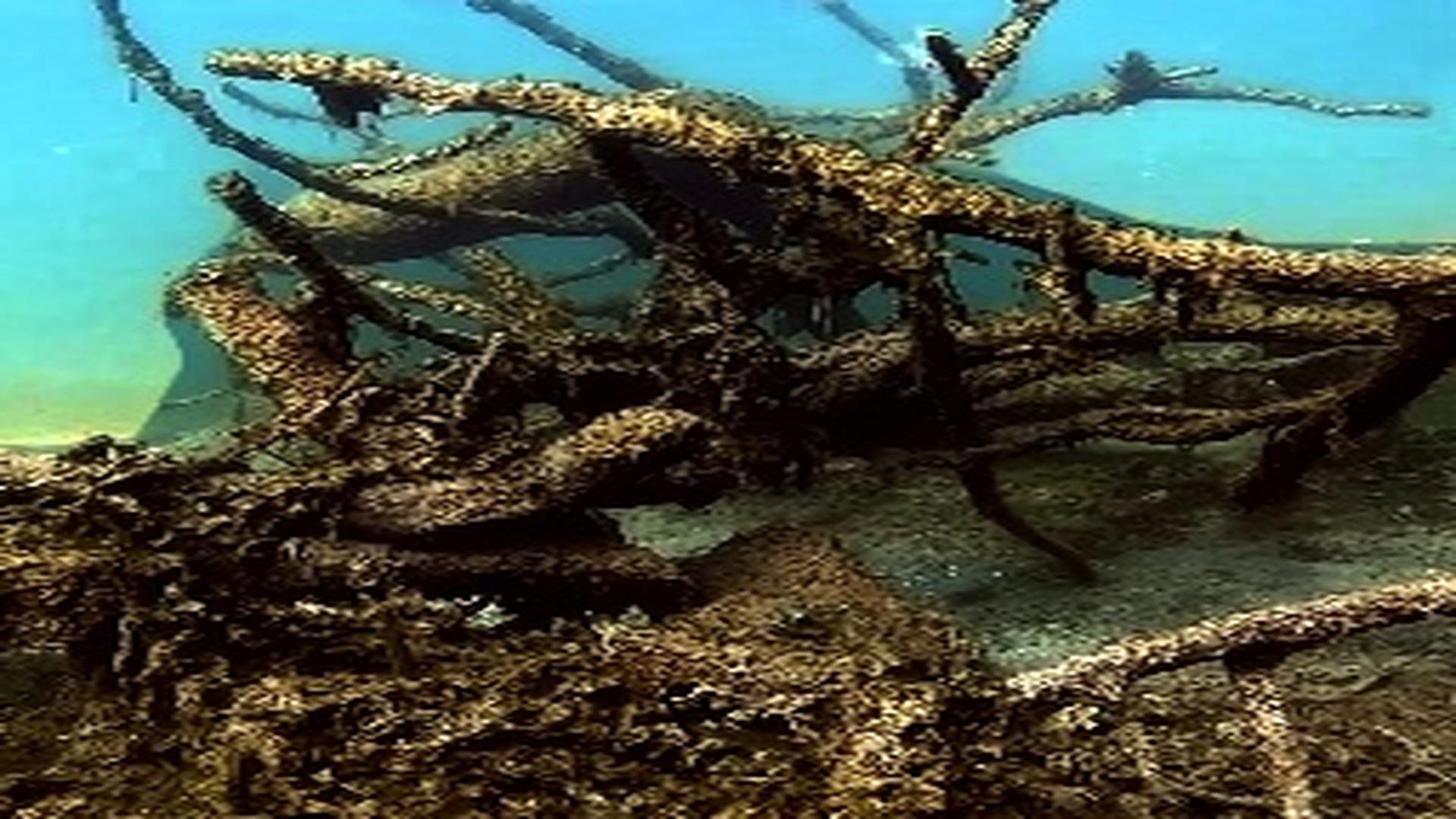} &
    \includegraphics[width=\linewidth]{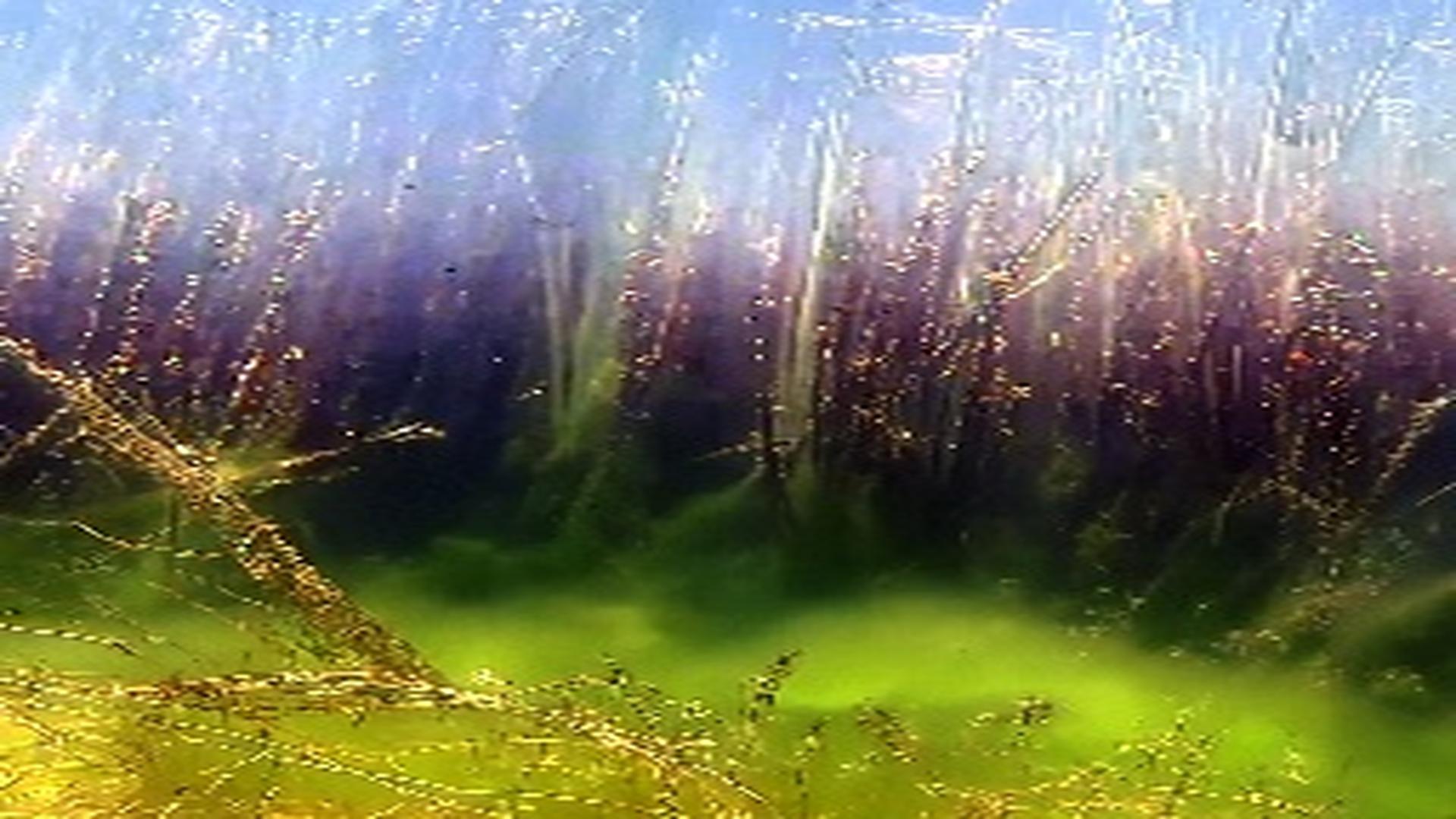} &
    \includegraphics[width=\linewidth]{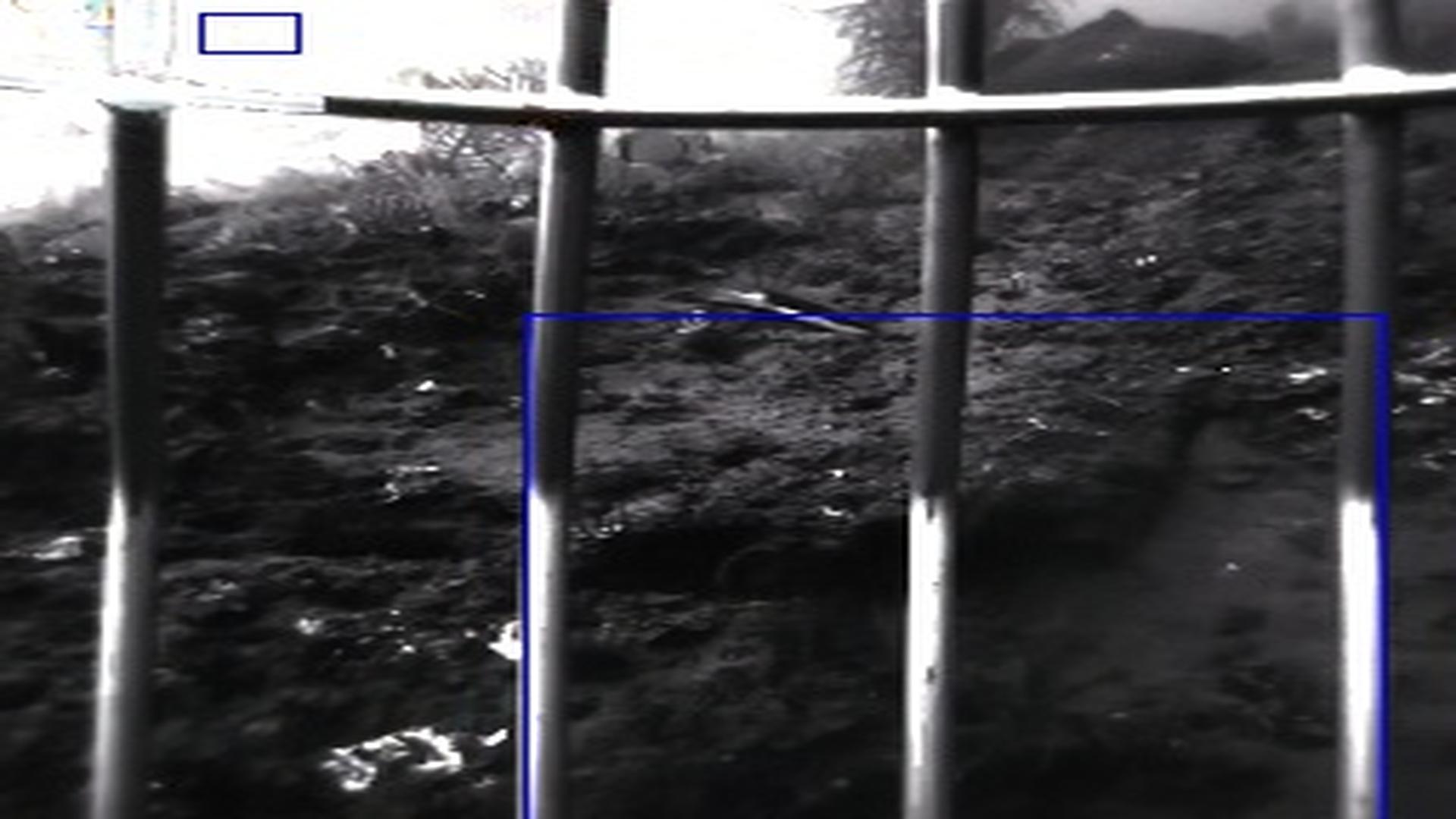} &
    \includegraphics[width=\linewidth]{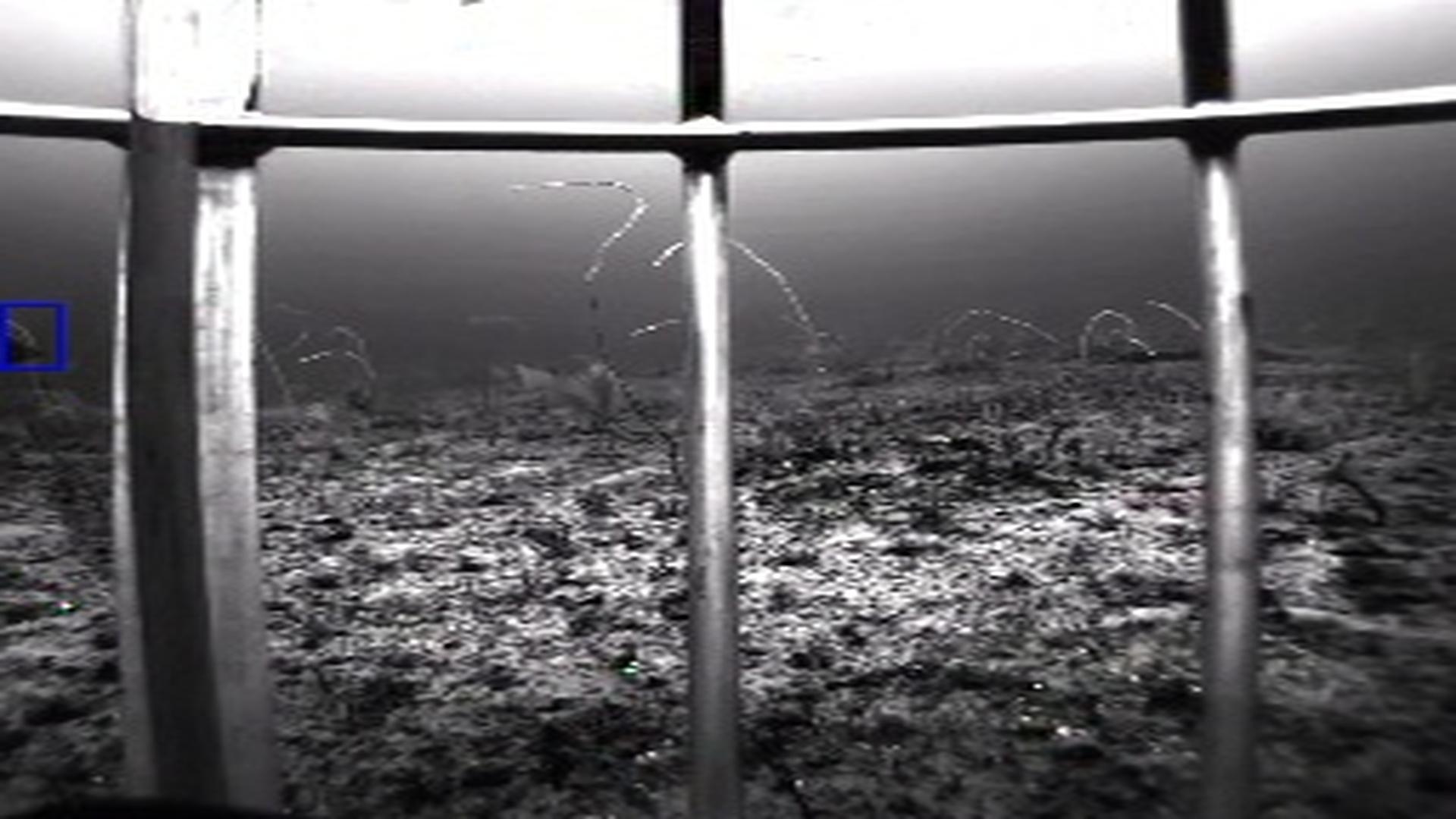} \\[-4pt]

    % Row 5: UDNet
    \scriptsize UDNet \cite{saleh2022adaptive} &
    \includegraphics[width=\linewidth]{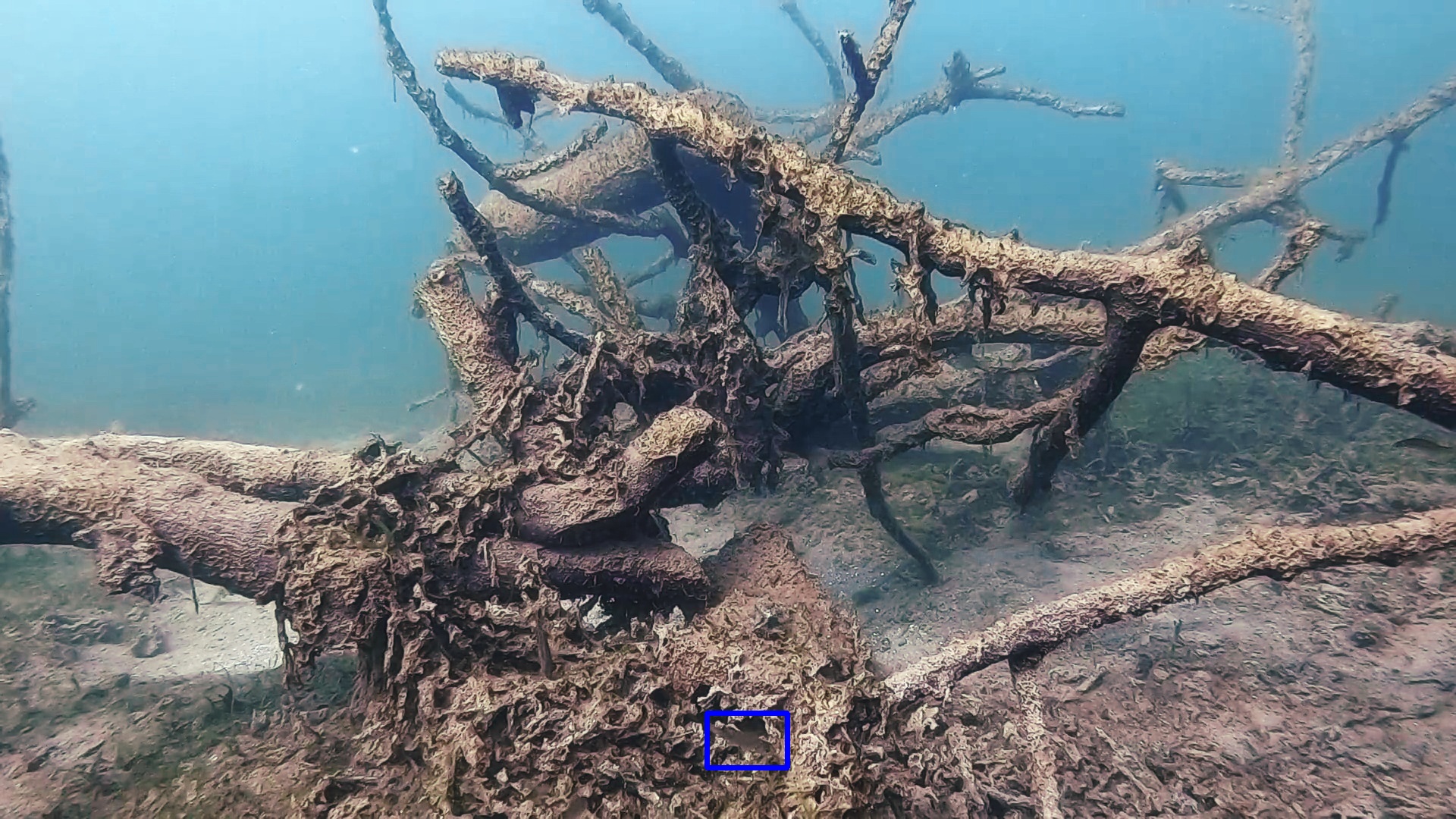} &
    \includegraphics[width=\linewidth]{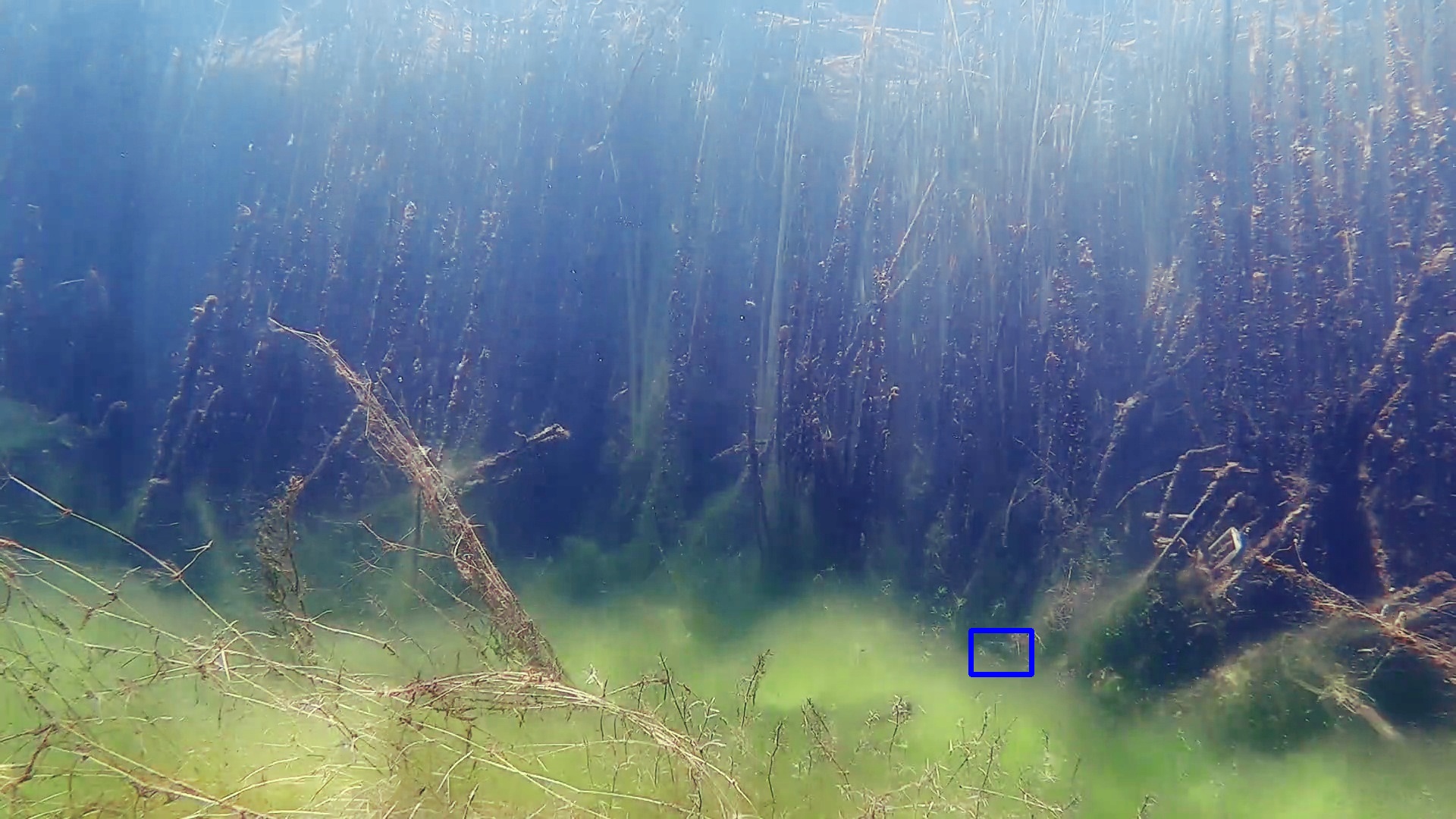} &
    \includegraphics[width=\linewidth]{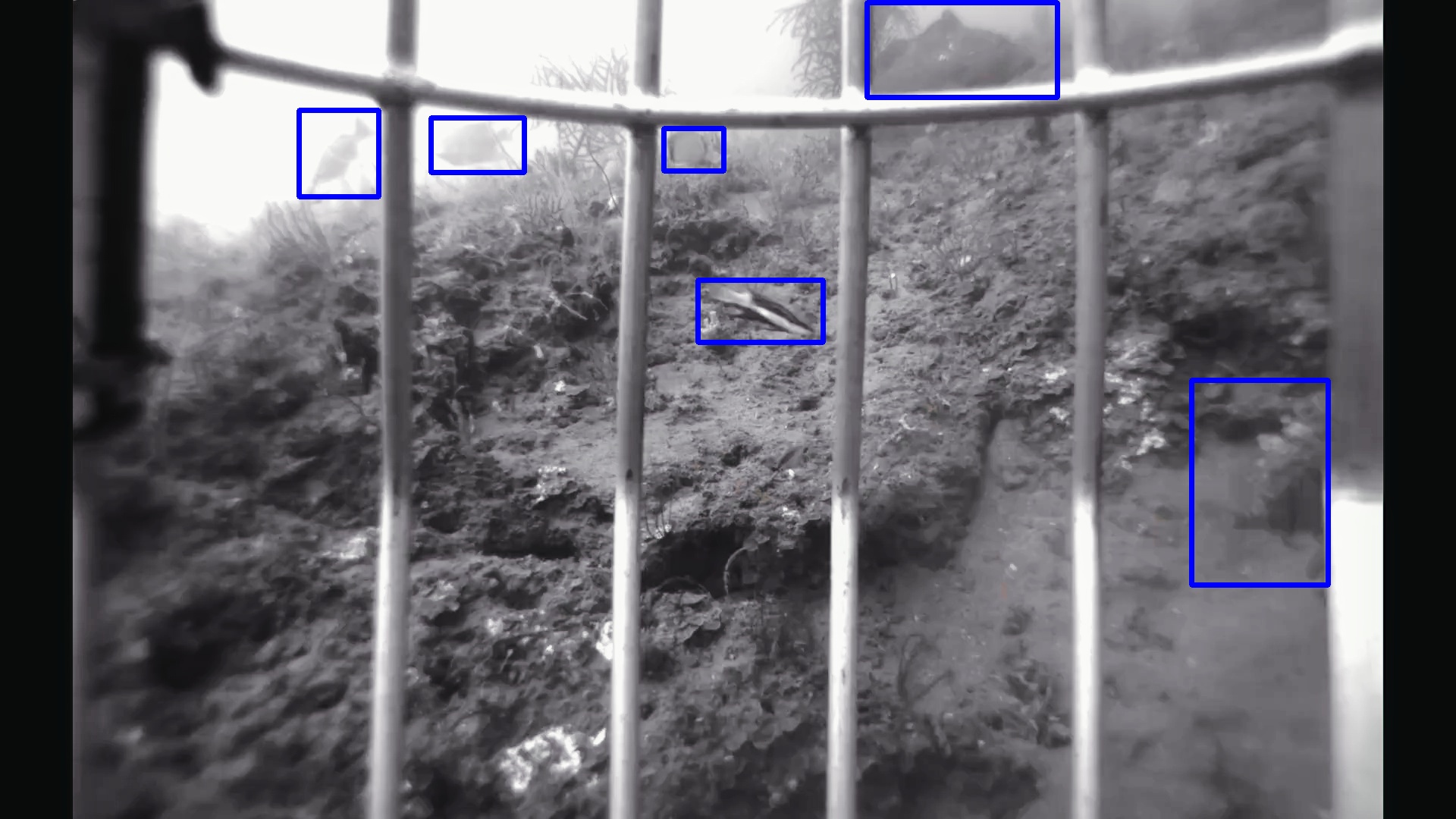} &
    \includegraphics[width=\linewidth]{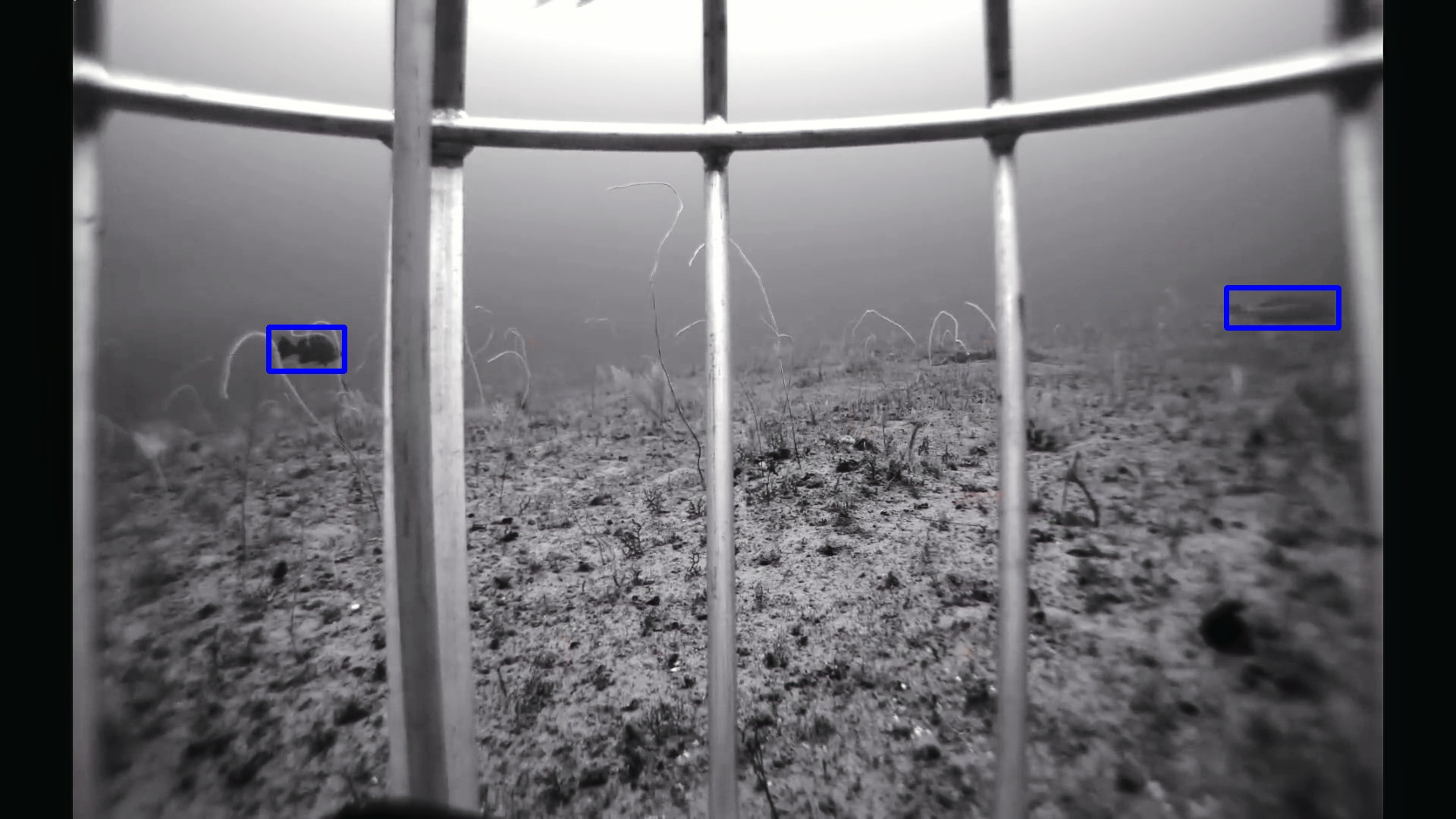} \\[-4pt]

    % Row 6: UDBE
    \scriptsize UDBE \cite{schein2025udbe} &
    \includegraphics[width=\linewidth]{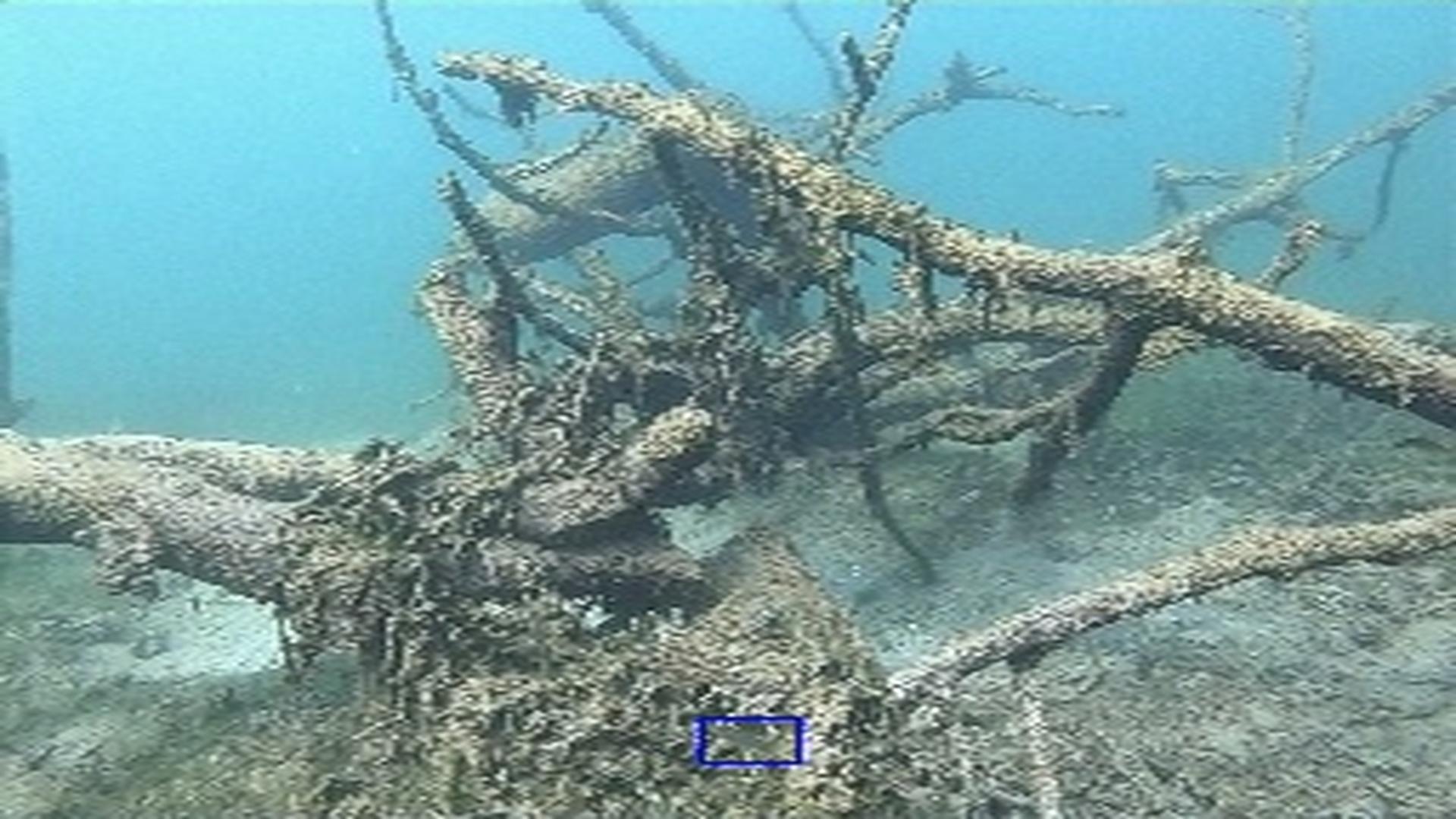} &
    \includegraphics[width=\linewidth]{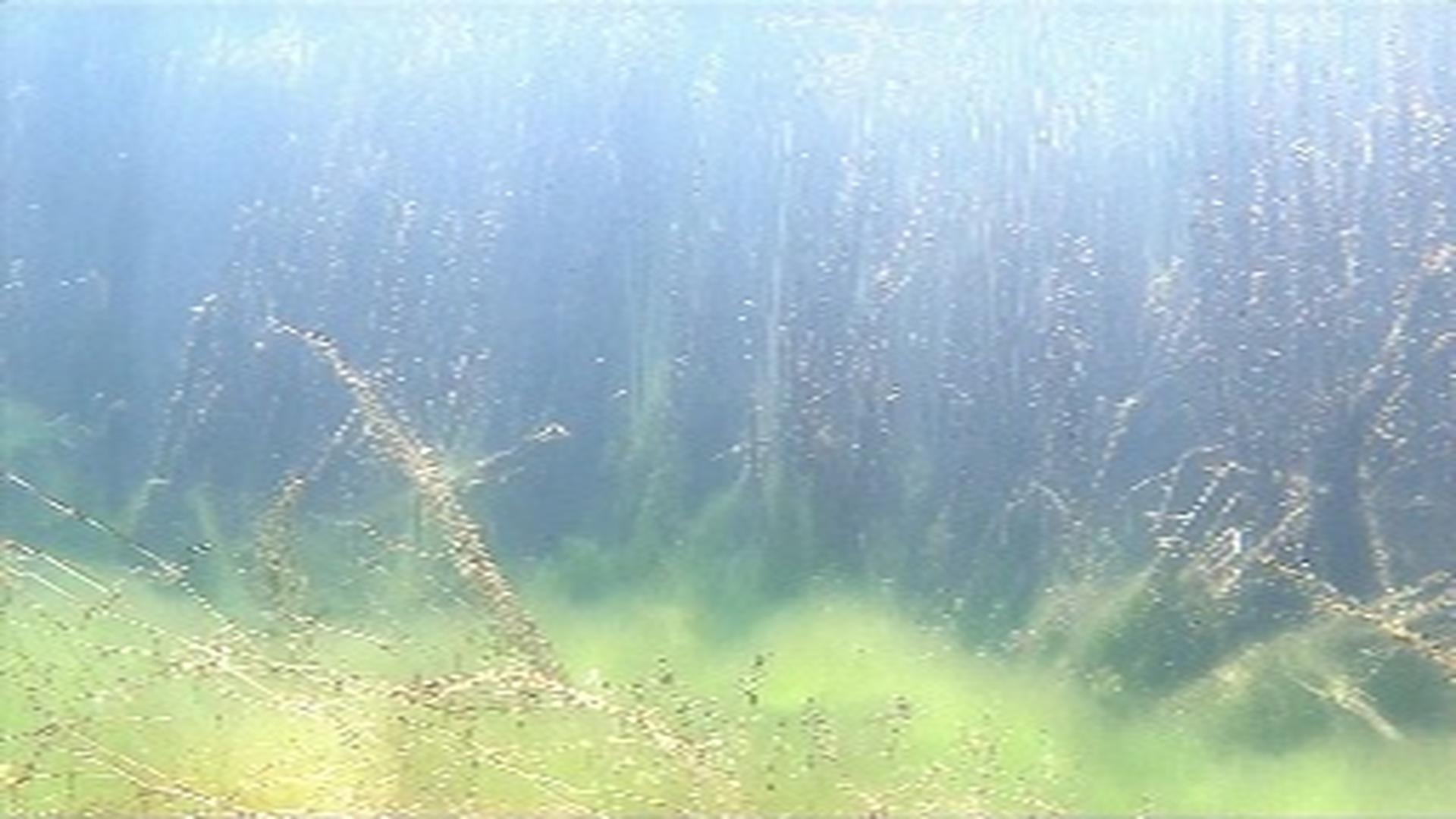} &
    \includegraphics[width=\linewidth]{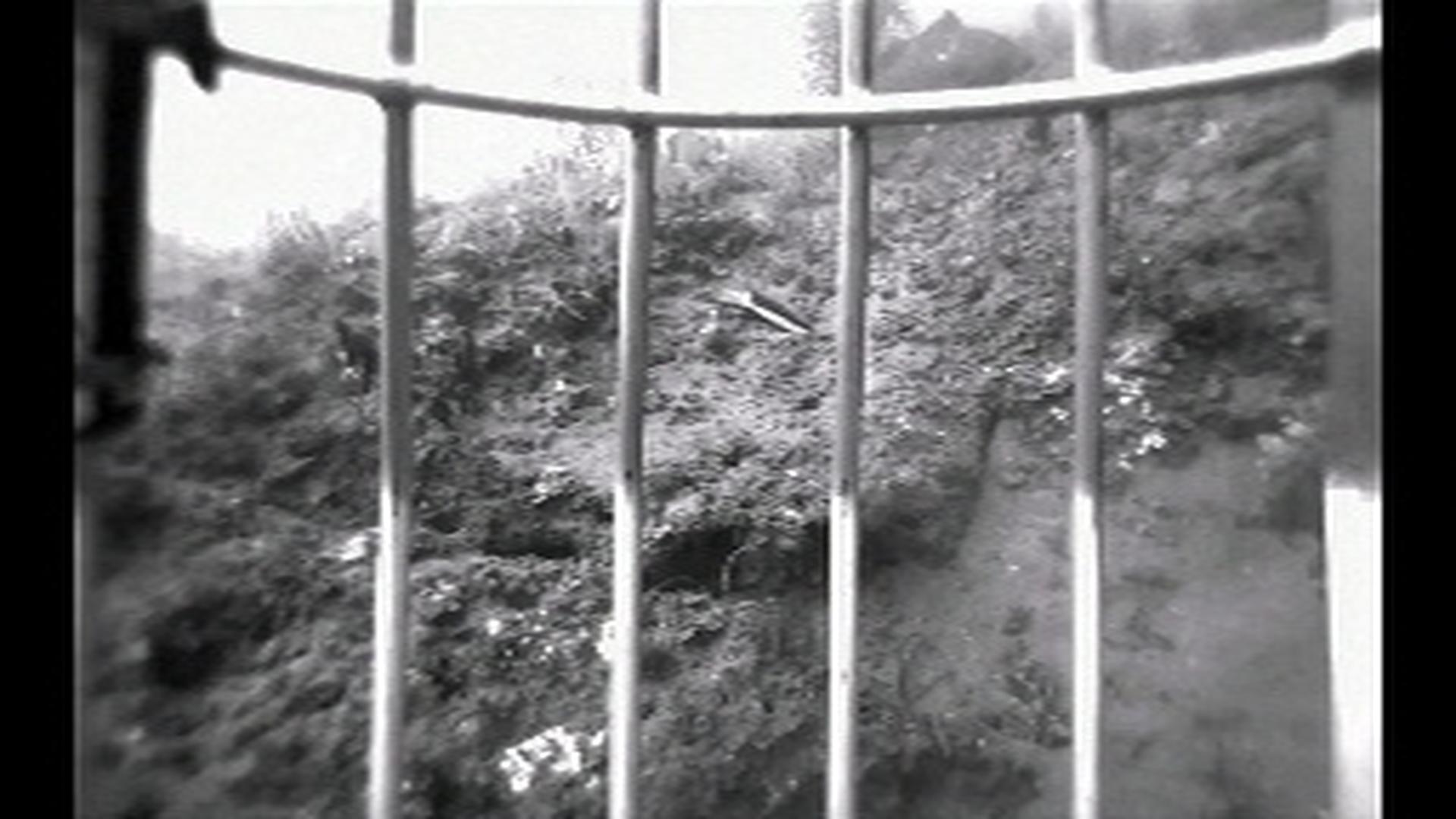} &
    \includegraphics[width=\linewidth]{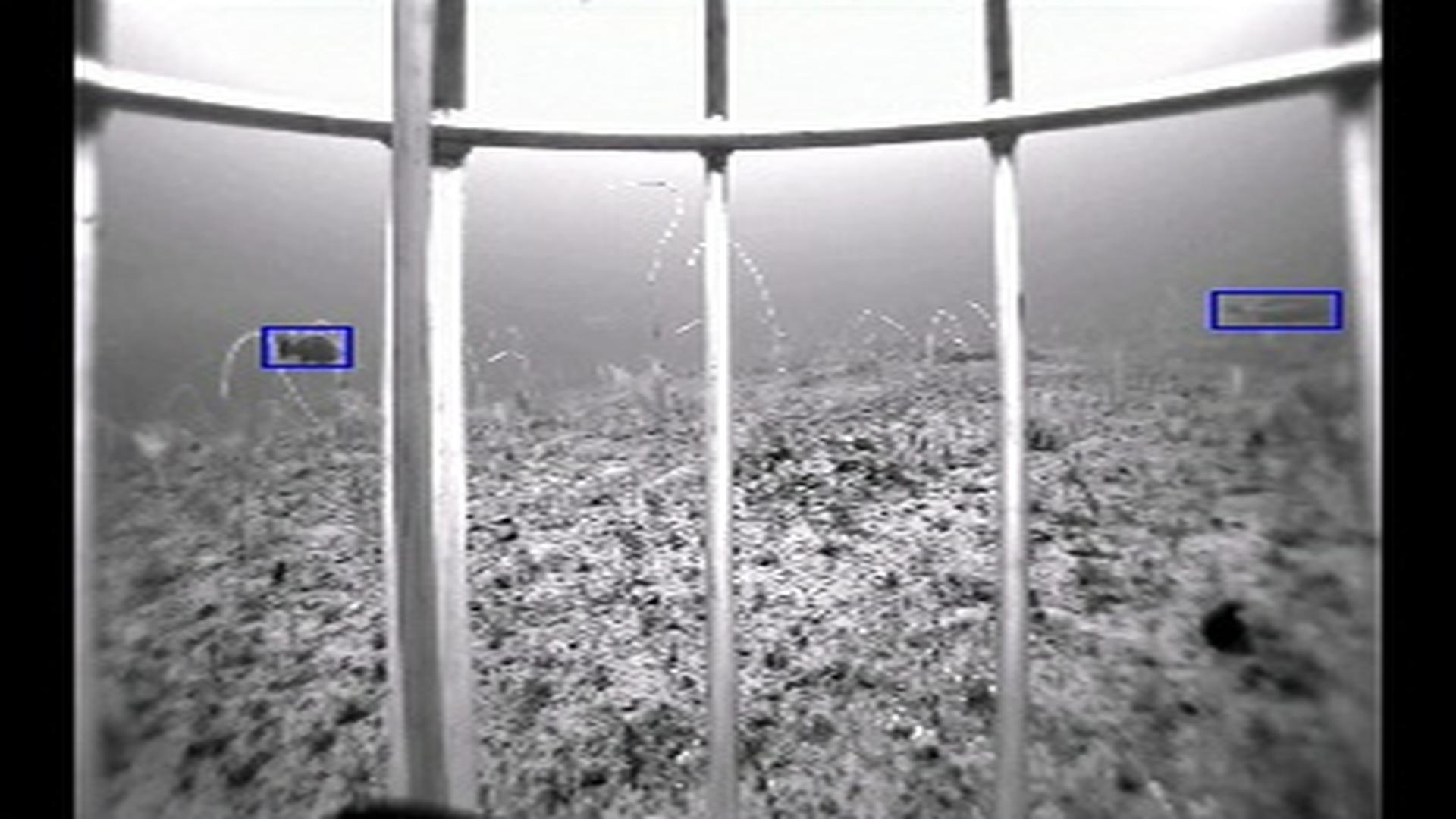} \\[-4pt]

    % Row 7: AquaFeat (Ours)
    \scriptsize \textbf{AquaFeat (Ours)} &
    \includegraphics[width=\linewidth]{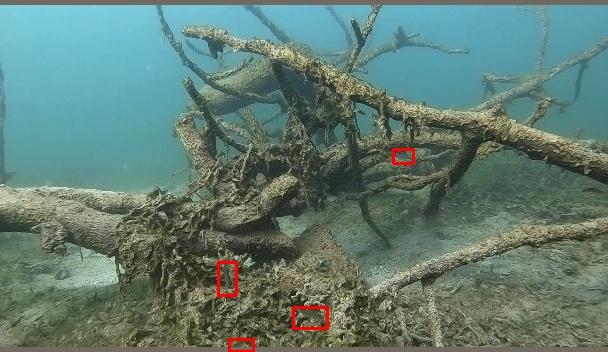} &
    \includegraphics[width=\linewidth]{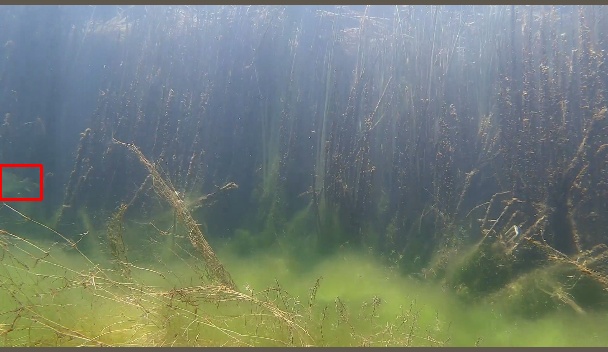} &
    \includegraphics[width=\linewidth]{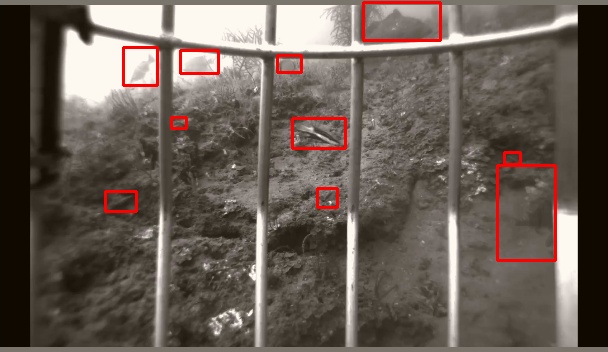} &
    \includegraphics[width=\linewidth]{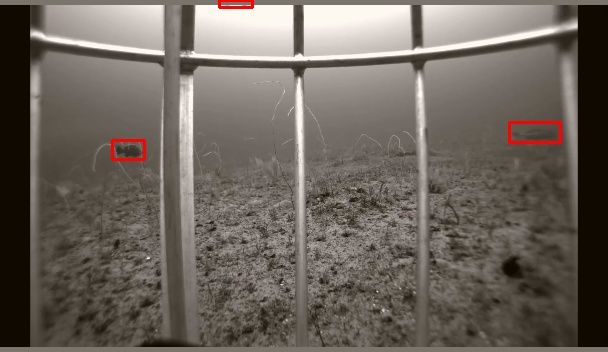} \\
    \end{tabular}
    \caption{Qualitative object detection comparison on FishTrack23 dataset. Each column shows a different scene. The rows display the ground truth (top), results from competing methods, and our proposed AquaFeat model (bottom).}
    \label{fig:qualitative}
\end{figure*}

In the qualitative analysis, we compared the object detection results of our proposed AquaFeat enhancement method with those of established visual enhancement techniques and the baseline YOLOv8m model. As illustrated in Fig. \ref{fig:qualitative}, AquaFeat consistently improved detection by enhancing image visibility and enabling more precise object localization. Most notably, as demonstrated in the scenario of the last column, AquaFeat was the only method capable of correctly detecting a partially occluded object at the image border, a task that all other techniques failed to accomplish. In the first column, both AquaFeat and AMSP-UOD missed one object compared to the ground truth (GT), while the original FeatEnhancer missed two. In the second column, several methods produced false positives, a pitfall avoided by AquaFeat and AMSP-UOD, which perfectly matched the GT. In the third column, AMSP-UOD identified two non-existent objects and failed to detect two present ones, whereas AquaFeat's results remained aligned with the GT. 

It is worth noting that the unsatisfactory performance of the Osmosis and UDBE methods may be attributed to their alteration of the original image dimensions, which can distort the scene and negatively impact detection. It should be noted that all methods receive the same input image, ensuring fair comparison conditions. For visual presentation purposes, all images from all methods were resized to uniform dimensions in the figures shown. Furthermore, UDBE introduced excessive noise in some images, hindering the detection task. Overall, the qualitative evaluation demonstrates that AquaFeat provides robust improvements in detection quality across various underwater contexts, showing superiority in complex scenarios where subtle details are critical for object identification.

    % Qualitative object detection comparison on challenging underwater images. The rows display the ground truth (top), results from competing methods, and our proposed AquaFeat model (bottom). 

\subsection{Quantitative Evaluation}

The validation of the AquaFeat method was conducted using fundamental object detection metrics. Precision measures the proportion of correct detections, while recall indicates the ability to identify all relevant objects present in the images \cite{manning2009introduction}. We employed the mean Average Precision (mAP) metric in two configurations: $mAP_{0.5}$, which requires a minimum Intersection over Union (IoU) of 50\% for a true positive, and $mAP_{0.5:0.95}$, which averages the mAP over IoU thresholds from 50\% to 95\%. Temporal performance was assessed using Frames Per Second (FPS). For precision, recall, and mAP, higher values denote better performance, whereas for FPS, higher values indicate superior temporal efficiency.

Table \ref{table_results_comparison} presents a comparison between the proposed approach and the reference algorithms, where the information in bold text refers to the highest score and the underlined one is the second best result. The proposed method excels in both Precision and Recall, achieving the highest scores among all compared techniques. The AquaFeat + YOLOv8m configuration obtained a Precision of $0.877$ and a Recall of $0.624$. These results underscore the robustness of our method, demonstrating a high capacity for correct detections (low false positive rate) and the ability to identify the majority of existing objects (low false negative rate). High precision is particularly crucial in underwater environments, where erroneous detections can lead to unnecessary or costly actions.

%A validação do método AquaFeat utilizou métricas fundamentais de detecção de objetos. A precisão representa a proporção de detecções corretas \cite{manning2009introduction}. O recall indica a capacidade de identificar todos os objetos presentes \cite{manning2009introduction}.

%Empregou-se a métrica mAP em duas configurações: $mAP_{0.5}$ considera detecções corretas com sobreposição mínima de 50\%, enquanto $mAP_{0.5:0.95}$ calcula a média das precisões com sobreposições variando de 50\% a 95\%. Avaliou-se também o desempenho temporal através da taxa FPS. Para precision, recall e mAP, valores superiores indicam melhor desempenho; para FPS, valores superiores representam melhor desempenho temporal.

%A Tabela \ref{table_results_comparison} apresenta a comparação entre a abordagem proposta e os algoritmos utilizados como referência. Pode-se observar que o método proposto se destaca nas métricas Precision e Recall, alcançando os maiores valores entre todos os métodos comparados. O AquaFeat + YOLOv8m obteve Precision de $0.877$ e Recall de $0.624$, indicando a robustez do método proposto, pois demonstra sua alta capacidade de realizar detecções corretas (baixa taxa de falsos positivos) e de identificar a maioria dos objetos presentes nas imagens (baixa taxa de falsos negativos). A alta precisão é particularmente importante em ambientes subaquáticos, onde detecções errôneas podem levar a ações desnecessárias ou dispendiosas.

\begin{table*}
\renewcommand{\arraystretch}{1.3}
\caption{Results Comparison}
\label{table_results_comparison}
\centering
\begin{tabular}{|c||c|c|c|c|c|c|}
\hline
\textbf{Method} & \textbf{mAP50$\uparrow$} & \textbf{mAP50-95$\uparrow$} & \textbf{Precision$\uparrow$} & \textbf{Recall$\uparrow$} & \textbf{FPS$\uparrow$} \\
\hline
YOLOv8m \cite{yoloV8} & 0.647 & 0.387 & 0.847 & 0.584 & \underline{321.54} \\
\hline
YOLOv10s \cite{yoloV8} & 0.592 & 0.325 & 0.777 & 0.549 & \textbf{700.18} \\
\hline
FeatEnhancer \cite{hashmi2023featenhancer} & 0.649 & 0.384 & 0.838 & 0.593 & 60.24 \\
%\hline
%CC+FeatEnhancer \cite{hashmi2023featenhancer} & 0.654 & 0.405 & 0.866 & 0.610 & 22.68ms & 44 \\
%\hline
%AquaFeat= CC+FeatEnhancer+specialconv+YOLOv8m \textbf{(Our)} & 0.658 & 0.401 & 0.871 & 0.601 & 23.18ms & 43 \\
%\hline
%AquaFeat \textbf{(Our)} & 0.677 & 0.421 & \textbf{0.877} & \textbf{0.624} & 23.15ms & 43 \\
%\hline
%AquaFeat= CC+FeatEnhancer+specialconv+YOLOv8m+NewSAFA \textbf{(Our)} & 0.675 & 0.425 & 0.893 & 0.617 & 24.36ms & 41 \\
\hline
%AMSP-UOD \cite{zhou2024amsp} & 0.732 & 0.462 & 0.846 & 0.599 & 24.2ms & 41 \\
AMSP-UOD \cite{zhou2024amsp} & \textbf{0.724} & \textbf{0.46} & \underline{0.866} & 0.578  & 41.84 \\
\hline
OSMOSIS+YOLOv8m \cite{nathan2024osmosis} & 0.023 & 0.005 & 0.133 & 0.005 & 0.0037 \\
\hline
UDNET+YOLOv8m \cite{saleh2022adaptive} & 0.566 & 0.336 & 0.810 & 0.505 & 15.92 \\
\hline
UDBE+YOLOv8m \cite{schein2025udbe} & 0.379 & 0.195 & 0.691 & 0.345 & 0.14  \\
\hline
AquaFeat \textbf{(Our)} + YOLOv8m & \underline{0.677} & \underline{0.421} & \textbf{0.877} & \textbf{0.624} & 46.51 \\
\hline
AquaFeat \textbf{(Our)} + YOLOv10s & 0.676 & \underline{0.421} & 0.859 & \underline{0.621} & 49.24 \\
\hline
\end{tabular}
\end{table*}

In terms of $mAP_{0.5}$ and $mAP_{0.5:0.95}$, the AMSP-UOD method \cite{zhou2024amsp} demonstrated the best performance with values of $0.724$ and $0.46$, respectively. Although our AquaFeat + YOLOv8m configuration did not surpass AMSP-UOD in these metrics, it achieved competitive results of $0.677$ for $mAP_{0.5}$ and $0.421$ for $mAP_{0.5:0.95}$, significantly outperforming all other methods. To test the reliability of our model with other detectors, we also trained it using YOLOv10s. The results (see Table \ref{table_results_comparison}) showed that AquaFeat + YOLOv10s improved the baseline YOLOv10s model across all key metrics: $mAP_{0.5}$ increased by 14.2\% (from $0.592$ to $0.676$), $mAP_{0.5:0.95}$ by 29.5\% (from $0.325$ to $0.421$), Precision by 10.6\% (from $0.777$ to $0.859$), and Recall by 13.1\% (from $0.549$ to $0.621$). While this configuration achieved a similar performance to AquaFeat + YOLOv8m, we used YOLOv8m as the primary baseline for its widespread adoption and balanced performance in the underwater object detection literature.

A key advantage of AquaFeat over AMSP-UOD lies in its versatility. While AMSP-UOD was developed for a specific application, AquaFeat can be integrated into various computer vision tasks. This flexibility allows for seamless updates to the detection model with newer versions, as demonstrated by the successful integration with YOLOv10s. Regarding temporal performance (FPS), the YOLOv10s was the fastest at 700.18 FPS. The AquaFeat + YOLOv8m and AquaFeat + YOLOv10s configurations achieved 46.51 FPS and 49.24 FPS, respectively. Although not as fast as the standalone YOLO models, AquaFeat provides adequate temporal performance, especially considering the substantial gains in precision and recall. Overall, traditional visual enhancement methods, when paired with YOLOv8m, failed to yield competitive results, demonstrating inferior performance compared to feature-based enhancement approaches, such as AquaFeat.

\section{Conclusion}

This work presented AquaFeat, an innovative plug-and-play module that enhances hierarchical features to optimize object detection in low-light underwater images. Experimental results demonstrated the method's superiority, outperforming the tested approaches in both Precision and Recall. The versatility of AquaFeat was validated by its successful integration into different architectures (YOLOv8m and YOLOv10s), establishing it as a flexible and high-impact solution. The study confirms that feature-based enhancement is more effective for underwater object detection than traditional visual enhancement methods. Future work will involve extending AquaFeat's application to object tracking in video and evaluating its effectiveness in other computer vision tasks, such as semantic segmentation and depth estimation.

\bibliographystyle{IEEEtran}
\bibliography{main}

\begin{thebibliography}{10}
\providecommand{\url}[1]{#1}
\csname url@samestyle\endcsname
\providecommand{\newblock}{\relax}
\providecommand{\bibinfo}[2]{#2}
\providecommand{\BIBentrySTDinterwordspacing}{\spaceskip=0pt\relax}
\providecommand{\BIBentryALTinterwordstretchfactor}{4}
\providecommand{\BIBentryALTinterwordspacing}{\spaceskip=\fontdimen2\font plus
\BIBentryALTinterwordstretchfactor\fontdimen3\font minus \fontdimen4\font\relax}
\providecommand{\BIBforeignlanguage}[2]{{%
\expandafter\ifx\csname l@#1\endcsname\relax
\typeout{** WARNING: IEEEtran.bst: No hyphenation pattern has been}%
\typeout{** loaded for the language `#1'. Using the pattern for}%
\typeout{** the default language instead.}%
\else
\language=\csname l@#1\endcsname
\fi
#2}}
\providecommand{\BIBdecl}{\relax}
\BIBdecl

\bibitem{ye2025yes}
S.~Ye, W.~Huang, W.~Liu, L.~Chen, X.~Wang, and X.~Zhong, ``{YES}: You should examine suspect cues for low-light object detection,'' \emph{Computer Vision and Image Understanding}, vol. 251, p. 104271, 2025.

\bibitem{chen2024underwater}
L.~Chen, Y.~Huang, J.~Dong, Q.~Xu, S.~Kwong, H.~Lu, H.~Lu, and C.~Li, ``Underwater object detection in the era of artificial intelligence: Current, challenge, and future,'' \emph{arXiv preprint arXiv:2410.05577}, 2024.

\bibitem{rasheed2025advancing}
M.~T. Rasheed, H.~Khan, J.~Wang, and Y.~Kang, ``Advancing low-light image enhancement through deep learning: A comprehensive experimental study,'' \emph{Knowledge-Based Systems}, p. 113827, 2025.

\bibitem{zhao2025deep}
Q.~Zhao, G.~Li, B.~He, and R.~Shen, ``Deep learning for low-light vision: A comprehensive survey,'' \emph{IEEE Transactions on Neural Networks and Learning Systems}, 2025.

\bibitem{colorcorrectionANDSpecialConv}
C.~Liu, X.~Shu, L.~Pan, J.~Shi, and B.~Han, ``Multiscale underwater image enhancement in {RGB} and {HSV} color spaces,'' \emph{IEEE Transactions on Instrumentation and Measurement}, vol.~72, pp. 1--14, 2023.

\bibitem{lai2025color}
Y.~L. Lai, T.~F. Ang, U.~A. Bhatti, C.~S. Ku, Q.~Han, and L.~Y. Por, ``Color correction methods for underwater image enhancement: A systematic literature review,'' \emph{PloS one}, vol.~20, no.~3, p. e0317306, 2025.

\bibitem{yoloV8}
\BIBentryALTinterwordspacing
Ultralytics, ``{YOLO}-v8,'' GitHub repository, 2023, accessed: 2025-05-02. [Online]. Available: \url{https://github.com/autogyro/yolo-V8}
\BIBentrySTDinterwordspacing

\bibitem{deepfish}
A.~Saleh, I.~H. Laradji, D.~A. Konovalov, M.~Bradley, D.~Vazquez, and M.~Sheaves, ``A realistic fish-habitat dataset to evaluate algorithms for underwater visual analysis,'' \emph{Scientific reports}, vol.~10, no.~1, p. 14671, 2020.

\bibitem{ozfish2020}
{Australian Institute Of Marine Science}, ``{Ozfish dataset - Machine learning dataset for baited remote underwater video stations},'' \url{https://github.com/open-AIMS/ozfish}, 2020, accessed: 2025-06-20.

\bibitem{dawkins2024fishtrack23}
M.~Dawkins, J.~Prior, B.~Lewis, R.~Faillettaz, T.~Banez, M.~Salvi, A.~Rollo, J.~Simon, M.~Campbell, M.~Lucero \emph{et~al.}, ``Fishtrack23: An ensemble underwater dataset for multi-object tracking,'' in \emph{Proceedings of the IEEE/CVF Winter Conference on Applications of Computer Vision}, 2024, pp. 7167--7176.

\bibitem{ji2023deep}
K.~Ji, W.~Lei, and W.~Zhang, ``A deep retinex network for underwater low-light image enhancement,'' \emph{Machine Vision and Applications}, vol.~34, no.~6, p. 122, 2023.

\bibitem{guo2025underwater}
L.~Guo, X.~Liu, D.~Ye, X.~He, J.~Xia, and W.~Song, ``Underwater object detection algorithm integrating image enhancement and deformable convolution,'' \emph{Ecological Informatics}, p. 103185, 2025.

\bibitem{islam2020fast}
M.~J. Islam, Y.~Xia, and J.~Sattar, ``Fast underwater image enhancement for improved visual perception,'' \emph{IEEE Robotics and Automation Letters}, vol.~5, no.~2, pp. 3227--3234, 2020.

\bibitem{liu2024lightweight}
T.~Liu, K.~Zhu, X.~Wang, W.~Song, and H.~Wang, ``Lightweight underwater image adaptive enhancement based on zero-reference parameter estimation network,'' \emph{Frontiers in Marine Science}, vol.~11, p. 1378817, 2024.

\bibitem{schein2025udbe}
T.~T. Schein, G.~P. de~Almeira, S.~L. Bri{\~a}o, R.~A. de~Bem, F.~G. de~Oliveira, and P.~L. Drews-Jr, ``{UDBE}: Unsupervised diffusion-based brightness enhancement in underwater images,'' \emph{arXiv preprint arXiv:2501.16211}, 2025.

\bibitem{nathan2024osmosis}
O.~B. Nathan, D.~Levy, T.~Treibitz, and D.~Rosenbaum, ``Osmosis: Rgbd diffusion prior for underwater image restoration,'' in \emph{European Conference on Computer Vision}.\hskip 1em plus 0.5em minus 0.4em\relax Springer, 2024, pp. 302--319.

\bibitem{saleh2022adaptive}
A.~Saleh, M.~Sheaves, D.~Jerry, and M.~R. Azghadi, ``Adaptive uncertainty distribution in deep learning for unsupervised underwater image enhancement,'' \emph{arXiv preprint arXiv:2212.08983}, 2022.

\bibitem{hashmi2023featenhancer}
K.~A. Hashmi, G.~Kallempudi, D.~Stricker, and M.~Z. Afzal, ``Featenhancer: Enhancing hierarchical features for object detection and beyond under low-light vision,'' in \emph{Proceedings of the IEEE/CVF International Conference on Computer Vision}, 2023, pp. 6725--6735.

\bibitem{li2025multi}
M.~Li, W.~Liu, C.~Shao, B.~Qin, A.~Tian, and H.~Yu, ``Multi-scale feature enhancement method for underwater object detection,'' \emph{Symmetry}, vol.~17, no.~1, p.~63, 2025.

\bibitem{ding2024lightweight}
J.~Ding, J.~Hu, J.~Lin, and X.~Zhang, ``Lightweight enhanced yolov8n underwater object detection network for low light environments,'' \emph{Scientific Reports}, vol.~14, no.~1, p. 27922, 2024.

\bibitem{zhou2024amsp}
J.~Zhou, Z.~He, K.-M. Lam, Y.~Wang, W.~Zhang, C.~Guo, and C.~Li, ``Amsp-uod: When vortex convolution and stochastic perturbation meet underwater object detection,'' in \emph{Proceedings of the AAAI Conference on Artificial Intelligence}, vol.~38, no.~7, 2024, pp. 7659--7667.

\bibitem{YOLOv10}
A.~Wang, H.~Chen, L.~Liu, K.~Chen, Z.~Lin, J.~Han \emph{et~al.}, ``Yolov10: Real-time end-to-end object detection,'' \emph{Advances in Neural Information Processing Systems}, vol.~37, pp. 107\,984--108\,011, 2024.

\bibitem{manning2009introduction}
C.~D. Manning, \emph{An introduction to information retrieval}, 2009.

\end{thebibliography}

\end{document}